\newcommand{\journaltitle}{arXiv preprint}
\newcommand{\bibstyleName}{elsarticle-harv}

\documentclass[final,5p,times,twocolumn,authoryear]{elsarticle}

\usepackage{amssymb}
\usepackage{amsmath}

\usepackage{booktabs}
\usepackage{siunitx} %
\usepackage{multirow} %
\usepackage{array} %
\usepackage{threeparttable} %
\usepackage{makecell} %
\usepackage{float}
\usepackage{placeins}
\makeatletter
\let\clarity@bodyaddtocurcol\@addtocurcol
\makeatother
\usepackage{stfloats}
\makeatletter
\let\clarity@appendixaddtocurcol\@addtocurcol
\let\@addtocurcol\clarity@bodyaddtocurcol
\newcommand{\enableAppendixBottomFloats}{\let\@addtocurcol\clarity@appendixaddtocurcol}
\makeatother
\providecommand{\appendixfloatbarrier}{\FloatBarrier}
\usepackage[colorlinks=true]{hyperref} %
\renewcommand{\theHtable}{\thesection.\arabic{table}}
\renewcommand{\theHfigure}{\thesection.\arabic{figure}}
\usepackage{xcolor}
\definecolor{refblue}{HTML}{0080ac}
\usepackage{tabularx}
\usepackage[T1]{fontenc}
\usepackage[utf8]{inputenc}
\usepackage{arydshln}
\usepackage{orcidlink}

\newif\ifarxivbuild
\arxivbuildtrue

\usepackage[capitalise]{cleveref}
\crefname{figure}{Fig.}{Figs.}
\Crefname{figure}{Fig.}{Figs.}
\crefname{table}{Tab.}{Tabs.}
\Crefname{table}{Tab.}{Tabs.}
\crefname{equation}{Eq.}{Eqs.}
\Crefname{equation}{Eq.}{Eqs.}
\crefname{section}{Sec.}{Secs.}
\Crefname{section}{Sec.}{Secs.}
\crefname{subsection}{Sec.}{Secs.}
\Crefname{subsection}{Sec.}{Secs.}
\crefname{subsubsection}{Sec.}{Secs.}
\Crefname{subsubsection}{Sec.}{Secs.}
\crefname{chapter}{Ch.}{Chs.}
\Crefname{chapter}{Ch.}{Chs.}
\crefname{algorithm}{Alg.}{Algs.}
\Crefname{algorithm}{Alg.}{Algs.}
\crefname{table*}{Tab.}{Tabs.}
\Crefname{table*}{Tab.}{Tabs.}
\crefname{figure*}{Fig.}{Figs.}
\Crefname{figure*}{Fig.}{Figs.}
\crefformat{appendix}{#2#1#3}
\Crefformat{appendix}{#2#1#3}
\crefmultiformat{appendix}{#2#1#3}{ and~#2#1#3}{, #2#1#3}{ and~#2#1#3}
\crefname{subappendix}{Appendix}{Appendices}
\Crefname{subappendix}{Appendix}{Appendices}

\journal{\journaltitle}

\makeatletter
\let\arxivOrigIncludegraphics\includegraphics

\newcommand{\arxiv@setfile}[1]{\def\arxiv@file{#1}}

\newcommand{\arxiv@figwidth}[2]{%
  \expandafter\def\csname arxiv@width@#1\endcsname{%
    \arxivOrigIncludegraphics[#2]}}
\arxiv@figwidth{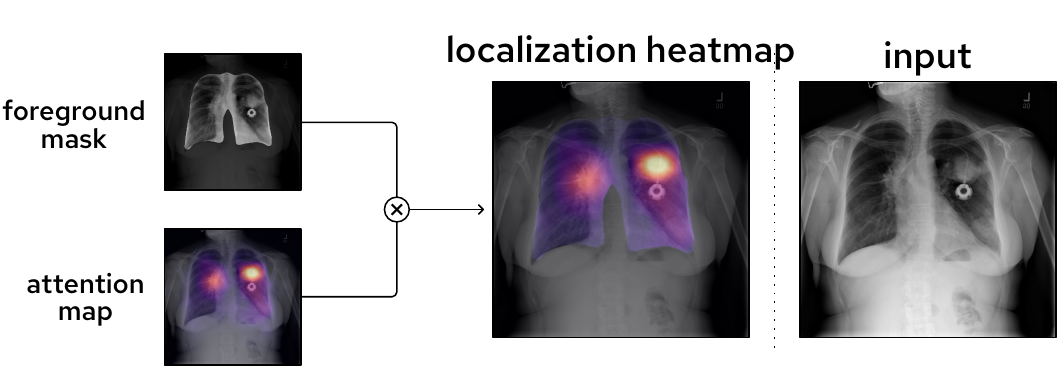}{width=\linewidth}
\arxiv@figwidth{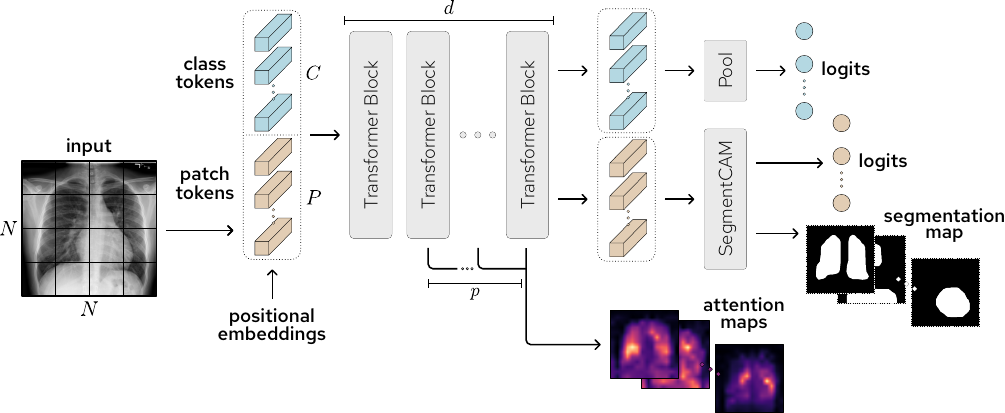}{width=0.95\linewidth}

\newcommand{\arxiv@wide}[1]{%
  \expandafter\def\csname arxiv@wide@#1\endcsname{1}}
\arxiv@wide{figure-1_main-architecture.pdf}
\arxiv@wide{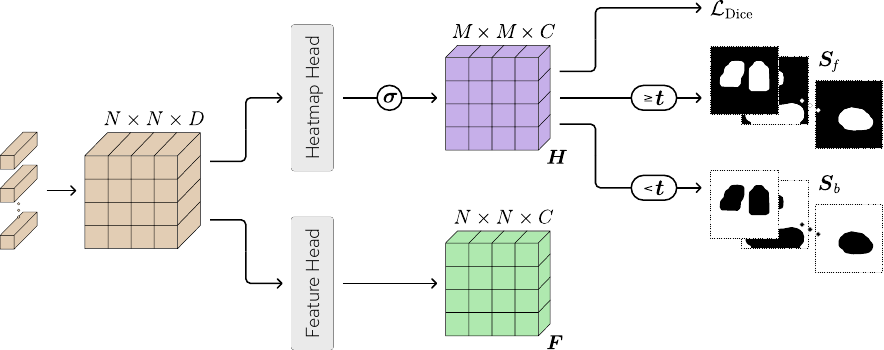}

\newcommand{\arxiv@figplace}[2]{%
  \expandafter\def\csname arxiv@place@#1\endcsname{#2}}
\arxiv@figplace{figure-13_transfer-overlays}{!t}

\newcommand{\arxiv@centerbox}[1]{\hbox to\linewidth{\hfil#1\hfil}}

\newif\ifarxiv@figurepending
\newif\ifarxiv@figurewide
\newif\ifarxiv@figureopen
\def\arxiv@figureplace{htbp}

\renewenvironment{figure}[1][htbp]{%
  \arxiv@figurependingtrue
  \arxiv@figurewidefalse
  \arxiv@figureopenfalse
  \def\arxiv@figureplace{#1}%
}{%
  \ifarxiv@figureopen
    \ifarxiv@figurewide\end@dblfloat\else\end@float\fi
  \else
    \arxiv@openfigure
    \ifarxiv@figurewide\end@dblfloat\else\end@float\fi
  \fi
}
\def\arxiv@openfigure{%
  \ifarxiv@figurewide
    \edef\arxiv@next{\noexpand\@dblfloat{figure}[\arxiv@figureplace]}%
  \else
    \edef\arxiv@next{\noexpand\@float{figure}[\arxiv@figureplace]}%
  \fi
  \arxiv@next
  \arxiv@figurependingfalse
  \arxiv@figureopentrue}

\renewcommand{\includegraphics}[2][]{%
  \ifarxiv@figurepending
    \ifcsname arxiv@wide@#2\endcsname\arxiv@figurewidetrue\fi
    \ifcsname arxiv@place@#2\endcsname
      \edef\arxiv@figureplace{\csname arxiv@place@#2\endcsname}%
    \fi
    \arxiv@openfigure
  \fi
  \arxiv@setfile{#2}%
  \ifcsname arxiv@width@#2\endcsname
    \arxiv@centerbox{\csname arxiv@width@#2\endcsname{\arxiv@file}}%
  \else
    \arxiv@centerbox{\arxivOrigIncludegraphics[#1]{\arxiv@file}}%
  \fi}
\makeatother

\usepackage{etoolbox}
\AtBeginEnvironment{table}{\setlength{\belowcaptionskip}{4pt}}
\AtBeginEnvironment{table*}{\setlength{\belowcaptionskip}{4pt}}

\renewcommand{\appendixfloatbarrier}{}

\makeatletter
\providecommand{\suppressappendixfloats}{%
  \suppressfloats[t]%
  \ifx\@dbldeferlist\@empty\global\@dbltopnum\z@\fi}
\makeatother

\makeatletter
\renewcommand{\FloatBarrier}{%
  \par
  \begingroup
    \let\@elt\relax
    \edef\arxiv@barriertest{\@deferlist\@dbldeferlist}%
    \ifx\arxiv@barriertest\@empty\else
      \ifnum\value{section}=3\else\ifnum\value{section}=10\else\clearpage\fi\fi
    \fi
  \endgroup}

\makeatletter
\newif\ifarxiv@jtabs
\let\arxiv@origsuppressappendixfloats\suppressappendixfloats
\renewcommand{\suppressappendixfloats}{%
  \ifnum\value{section}=1
  \begin{table}[htb]
  \centering
  \caption{Scalar loss weights selected by the tuning search to maximize the validation UDP. The same values are used in all experiments.}
  \label{tab:loss_weights}
  \begin{tabular}{l r}
  \toprule
  Loss Weight & {Value} \\
  \midrule
  $\alpha_C$ & 2.2674 \\
  $\alpha_{\hspace{-0.08em} P}$ & 2.4678 \\
  $\alpha_{\negthinspace f}$ & 3.9670 \\
  $\beta$ & 0.4377 \\
  $\gamma$ & 0.1427 \\
  \bottomrule
  \end{tabular}
  \end{table}
  \fi
  \ifnum\value{section}=10
    \ifarxiv@jtabs\else
      \arxiv@jtabstrue
      \begin{table*}[!t]
      \centering
      \caption{VinDr-CXR per-finding image-level IoU accuracy by IoU threshold $T$ (test split, six box findings, six model settings). Each cell is the fraction of that finding's positive test images whose predicted boxes contain at least one box with IoU $\geq T$; $n$ is the number of test positives. The macro row is the unweighted mean over the six findings and matches the VinDr-CXR macro of the per-threshold table to three decimals. Point estimates; per-finding intervals are not computed. The highest value for each finding and threshold is shown in bold.}
      \label{tab:transfer_vindr_per_finding}
      \footnotesize
      \setlength{\tabcolsep}{1pt}
      \begin{tabular*}{0.495\textwidth}[t]{@{\extracolsep{\fill}} l l r r r r r@{}}
      \toprule
      Finding & Model & $T{=}0.1$ & $T{=}0.2$ & $T{=}0.3$ & $T{=}0.4$ & $T{=}0.5$ \\
      \midrule
      \multirow{6}{*}{\shortstack[l]{Atelectasis\\$n{=}86$}} & CLARiTy NIH-cal., 224 & 0.616 & 0.419 & 0.279 & 0.163 & 0.081 \\
       & CLARiTy NIH-cal., 512 & \bfseries 0.674 & \bfseries 0.488 & \bfseries 0.372 & \bfseries 0.279 & \bfseries 0.186 \\
       & CLARiTy Target-cal., 224 & 0.512 & 0.384 & 0.267 & 0.186 & 0.081 \\
       & CLARiTy Target-cal., 512 & 0.500 & 0.279 & 0.198 & 0.140 & 0.105 \\
       & RAD-DINO 518 & 0.512 & 0.302 & 0.186 & 0.081 & 0.035 \\
       & TXV-NIH 224 & 0.256 & 0.070 & 0.058 & 0.000 & 0.000 \\
      \midrule
      \multirow{6}{*}{\shortstack[l]{Cardiomegaly\\$n{=}309$}} & CLARiTy NIH-cal., 224 & \bfseries 1.000 & \bfseries 1.000 & \bfseries 1.000 & 0.916 & 0.573 \\
       & CLARiTy NIH-cal., 512 & \bfseries 1.000 & \bfseries 1.000 & \bfseries 1.000 & 0.909 & 0.586 \\
       & CLARiTy Target-cal., 224 & \bfseries 1.000 & \bfseries 1.000 & 0.997 & 0.971 & 0.864 \\
       & CLARiTy Target-cal., 512 & 0.997 & 0.997 & 0.997 & \bfseries 0.974 & \bfseries 0.926 \\
       & RAD-DINO 518 & 0.987 & 0.858 & 0.430 & 0.091 & 0.019 \\
       & TXV-NIH 224 & 0.942 & 0.599 & 0.139 & 0.023 & 0.006 \\
      \midrule
      \multirow{6}{*}{\shortstack[l]{Pleural\\effusion\\$n{=}111$}} & CLARiTy NIH-cal., 224 & 0.586 & 0.252 & 0.117 & 0.054 & 0.018 \\
       & CLARiTy NIH-cal., 512 & 0.486 & 0.198 & 0.054 & 0.045 & 0.018 \\
       & CLARiTy Target-cal., 224 & 0.577 & 0.324 & 0.135 & 0.081 & 0.027 \\
       & CLARiTy Target-cal., 512 & 0.604 & 0.351 & 0.198 & 0.117 & 0.054 \\
       & RAD-DINO 518 & \bfseries 0.784 & \bfseries 0.559 & \bfseries 0.333 & \bfseries 0.162 & \bfseries 0.072 \\
       & TXV-NIH 224 & 0.396 & 0.153 & 0.081 & 0.063 & 0.036 \\
      \midrule
      \multirow{6}{*}{\shortstack[l]{Infiltration\\$n{=}58$}} & CLARiTy NIH-cal., 224 & 0.690 & 0.552 & 0.500 & \bfseries 0.397 & 0.207 \\
       & CLARiTy NIH-cal., 512 & 0.828 & 0.690 & \bfseries 0.517 & 0.362 & 0.259 \\
       & CLARiTy Target-cal., 224 & 0.741 & 0.552 & 0.448 & 0.379 & \bfseries 0.293 \\
       & CLARiTy Target-cal., 512 & \bfseries 0.845 & \bfseries 0.724 & \bfseries 0.517 & 0.362 & 0.241 \\
       & RAD-DINO 518 & 0.810 & 0.655 & 0.431 & 0.190 & 0.155 \\
       & TXV-NIH 224 & 0.586 & 0.379 & 0.259 & 0.138 & 0.052 \\
      \bottomrule
      \end{tabular*}\hfill
      \begin{tabular*}{0.495\textwidth}[t]{@{\extracolsep{\fill}} l l r r r r r@{}}
      \toprule
      Finding & Model & $T{=}0.1$ & $T{=}0.2$ & $T{=}0.3$ & $T{=}0.4$ & $T{=}0.5$ \\
      \midrule
      \multirow{6}{*}{\shortstack[l]{Nodule/Mass\\$n{=}176$}} & CLARiTy NIH-cal., 224 & 0.597 & 0.472 & 0.381 & 0.295 & 0.199 \\
       & CLARiTy NIH-cal., 512 & 0.784 & 0.693 & \bfseries 0.597 & \bfseries 0.494 & \bfseries 0.386 \\
       & CLARiTy Target-cal., 224 & 0.608 & 0.466 & 0.381 & 0.267 & 0.188 \\
       & CLARiTy Target-cal., 512 & \bfseries 0.795 & \bfseries 0.699 & 0.591 & 0.449 & 0.330 \\
       & RAD-DINO 518 & 0.665 & 0.534 & 0.364 & 0.261 & 0.159 \\
       & TXV-NIH 224 & 0.318 & 0.256 & 0.136 & 0.068 & 0.034 \\
      \midrule
      \multirow{6}{*}{\shortstack[l]{Pneumothorax\\$n{=}18$}} & CLARiTy NIH-cal., 224 & 0.611 & 0.389 & 0.167 & 0.111 & \bfseries 0.056 \\
       & CLARiTy NIH-cal., 512 & 0.722 & 0.333 & 0.167 & 0.056 & \bfseries 0.056 \\
       & CLARiTy Target-cal., 224 & 0.667 & 0.389 & 0.167 & 0.111 & \bfseries 0.056 \\
       & CLARiTy Target-cal., 512 & 0.500 & 0.222 & 0.056 & 0.056 & \bfseries 0.056 \\
       & RAD-DINO 518 & \bfseries 0.889 & \bfseries 0.778 & 0.389 & 0.111 & 0.000 \\
       & TXV-NIH 224 & 0.722 & 0.611 & \bfseries 0.500 & \bfseries 0.278 & \bfseries 0.056 \\
      \midrule
      \multirow{6}{*}{\shortstack[l]{macro\\(6 findings)}} & CLARiTy NIH-cal., 224 & 0.683 & 0.514 & 0.407 & 0.323 & 0.189 \\
       & CLARiTy NIH-cal., 512 & 0.749 & 0.567 & \bfseries 0.451 & \bfseries 0.358 & 0.248 \\
       & CLARiTy Target-cal., 224 & 0.684 & 0.519 & 0.399 & 0.333 & 0.251 \\
       & CLARiTy Target-cal., 512 & 0.707 & 0.545 & 0.426 & 0.350 & \bfseries 0.285 \\
       & RAD-DINO 518 & \bfseries 0.774 & \bfseries 0.614 & 0.356 & 0.149 & 0.073 \\
       & TXV-NIH 224 & 0.537 & 0.345 & 0.196 & 0.095 & 0.031 \\
      \bottomrule
      \end{tabular*}
      \end{table*}
    \fi
  \fi
  \arxiv@origsuppressappendixfloats}
\makeatother

\begin{document}

\hypersetup{
    linkcolor=refblue,
    citecolor=refblue,
    urlcolor=refblue
}

\begin{frontmatter}

\title{CLARiTy: A Vision Transformer for Multi-Label Classification and Weakly-Supervised Localization of Chest X-ray Pathologies} %

\author{John M. Statheros\corref{corres_label}\,\orcidlink{0009-0006-3675-2169}}
\ead{johnmstatheros@gmail.com}
\cortext[corres_label]{Corresponding author.}
\author{Hairong Wang\,\orcidlink{0000-0001-8770-5916}}
\ead{hairong.bau@wits.ac.za}
\author{Richard Klein\,\orcidlink{0000-0003-0783-2072}}
\ead{richard.klein@wits.ac.za}

\affiliation{organization={School of Computer Science and Applied Mathematics, University of the Witwatersrand},
            addressline={1 Jan Smuts Avenue},
            city={Johannesburg},
            postcode={2000},
            state={Gauteng},
            country={South Africa}}

\begin{abstract}
Chest X-ray interpretation requires multi-label classification and spatial localization, yet region-level annotations are scarce, so weakly supervised localization involves training on image-level labels to produce class heatmaps. The step that converts heatmaps into boxes is rarely calibrated, that is, fitted on validation images, and localization results seldom leave the training dataset. We present CLARiTy (Class Localizing and Attention Refining Image Transformer), a vision transformer whose class tokens yield class-specific attention maps without gradient class activation mapping. A SegmentCAM head trained on anatomical region masks, not pathology masks, suppresses background. OneBox then converts the heatmaps into boxes, with parameters fitted per finding and resolution. On the official NIH split, CLARiTy outperforms the compared methods at matched resolution in localization and remains competitive in classification. Against published ThoraX-PriorNet values at matched 224 px resolution, the relative gain in macro intersection-over-union (IoU) accuracy at IoU 0.5 is 50.7\%. The largest relative gain is on nodules: at that threshold the 224-pixel model reaches 0.154, where no compared method exceeds 0.013. Without training any weights on RSNA Pneumonia or VinDr-CXR, we refit only the OneBox parameters on target validation images, leaving the model and heatmaps unchanged. Localization is competitive with RAD-DINO and ahead of TorchXRayVision DenseNet121-NIH at every threshold. Fitting the box extractor on validation images, freezing it before testing and reporting localization across thresholds gives a protocol that carries to new corpora without retraining the model. Code is available at \url{https://github.com/jstat17/clarity-cxr}.
\end{abstract}


\end{frontmatter}

\section{Introduction}
Chest X-rays can diagnose many thoracic diseases, including COVID-19, pneumonia and tuberculosis~\citep{WHO2022}, and radiologists report not only which disease is present but also where in the image it lies. A classifier trained on image-level labels can learn the disease without producing its location, so weakly supervised localization reads boxes out of its heatmaps~\citep{Wang2017-NIH}. The heatmaps usually come from class activation mapping (CAM)~\citep{Zhou2016-CAM} or its gradient-weighted variant, Grad-CAM~\citep{Selvaraju2017-GradCAM}. Two properties make the conversion hard: the maps are coarse, computed on a feature grid much smaller than the image, and many findings are small, so a nodule or a mass covers only a few cells~\citep{Sedai2018}.

A predicted box counts as correct when its overlap with the annotated box, measured as the ratio of their intersection to their union, reaches a threshold; this ratio is the IoU, so localization depends on the threshold. A method's IoU accuracy is the share of annotated boxes for which some predicted box reaches that threshold. Currently the heatmap-to-box threshold is fixed by hand, and results seldom leave their dataset. Choosing the heatmap-to-box parameters on separate validation images~\citep{Choe2020-WSOL-Eval} and reporting accuracy across thresholds~\citep{Wang2017-NIH} are the established remedies. Sweeping the IoU threshold shows what a method achieves as the criterion tightens: a loose threshold is met by any overlap, whereas a strict one is met only by a box that fits the finding.

Vision transformers (ViTs) offer a second route through self-attention, but maps obtained by fusing attention across layers are sparse and unfocused, and in medical images localize less reliably than CAM~\citep{Barekatain2025}. Standalone ViT self-attention is therefore seldom used for chest X-ray localization, where hybrid convolutional-transformer models reading out CAM are more common.

We take the standalone route. CLARiTy is a vision transformer for multi-label classification and weakly supervised localization. To our knowledge, it is the first standalone ViT for weakly supervised box localization in chest X-rays to combine one class token per pathology (an extra learned token that represents that finding), SegmentCAM and attention pooling. Its localization map is the class-to-patch attention of the class tokens averaged over the last few transformer blocks, with no gradient CAM. A SegmentCAM head, supervised only by anatomical region masks (lungs, heart, mediastinum, diaphragm), predicts a foreground mask at image resolution and suppresses background without boxes or pathology masks. Ablations that remove each component in turn show that SegmentCAM improves localization on validation images (\cref{sec:ablation-study}). Classification passes the class tokens through attention pooling, which learns per-pathology weights over the embedding dimensions. Finally, OneBox turns each heatmap into boxes, with parameters fitted per finding and resolution on validation images rather than fixed by hand~(\cref{sec:onebox}). Together these components lead to the contributions below.

The contributions are:
\begin{itemize}
    \item CLARiTy, a standalone vision transformer that reads class-specific heatmaps from the attention of one class token per pathology. SegmentCAM background suppression trained on anatomical region masks, attention pooling and self-supervised (DINO) initialization each improve validation localization in ablations, attention pooling at a small cost in classification (\cref{sec:ablation-study}).
    \item OneBox, a heatmap-to-box extractor whose parameters are fitted per finding and resolution on validation images rather than fixed in advance. On the same maps, the uncapped fixed-threshold rule of Wang et al.~\citep{Wang2017-NIH} reaches a higher IoU accuracy than OneBox only at loose thresholds, and only by emitting 2.4 to 3.5 times as many boxes. With one box per instance, OneBox has the higher IoU accuracy at every threshold~(\cref{appx:onebox-robustness}).
    \item The highest macro IoU accuracy among the compared methods on the official NIH split (the patient-wise partition released with ChestX-ray14) at matched resolution (the same input size on both sides) at every reported threshold. At IoU 0.5 it reaches 0.303 at 224 px and 0.295 at 512 px, against the published 0.201 and 0.217 of ThoraX-PriorNet~\citep{Hossain2024-ThoraX-PriorNet}, a relative gain of 50.7\% at 224 px. The gain is largest on the smallest findings: on nodules the 224-pixel model reaches 0.154, where no compared method exceeds 0.013. Classification (macro AUC 0.808 at 224 px) is competitive with, not above, the best official-split models~(\cref{sec:results}).
    \item A cross-dataset study with the NIH weights frozen, on RSNA Pneumonia~\citep{Shih2019-RSNA} and VinDr-CXR~\citep{Nguyen2022-VinDrCXR}. Refitting only the OneBox parameters on each target's validation images raises image-level IoU accuracy (the share of a finding's positive images with a box reaching the threshold) at IoU 0.5 at both resolutions on both corpora. After refitting, CLARiTy is competitive with RAD-DINO~\citep{PerezGarcia2025-RADDINO}, a self-supervised vision transformer whose probe is fitted on target image labels. It is ahead from IoU 0.2 on RSNA and from IoU 0.3 on VinDr-CXR, where RAD-DINO leads at the loosest thresholds.
\end{itemize}
\section{Related work}\label{sec:related-work}
\subsection{Deep learning for multi-label chest X-ray pathology classification}
Convolutional neural networks (CNNs) have been the predominant architecture for multi-label CXR classification~\citep{Baltruschat2019}, and two of the key public corpora are NIH ChestX-ray14 (112,120 images, 14 labels)~\citep{Wang2017-NIH} and VinDr-CXR (18,000 images with boxes)~\citep{Nguyen2022-VinDrCXR}. Accuracy depends on the split: a random patient-wise split raised the mean AUC by 3.4\% absolute and an image-wise split by 10.8\% absolute~\citep{Gundel2019,Wang2019-Thorax-Net}, so comparisons without a common split can be inconclusive~\citep{Baltruschat2019}.

\subsection{Weakly-supervised localization in chest X-ray image analysis}
Weakly supervised localization derives boxes or heatmaps from image-level labels. CAM~\citep{Zhou2016-CAM} is its cornerstone: gradient-free variants aggregate activations, whereas Grad-CAM weights them by gradients and is the standard explainability baseline~\citep{Selvaraju2017-GradCAM}. CXR studies adopt Grad-CAM~\citep{Wang2019-Thorax-Net}, but its heatmaps are often large and regular-shaped, which costs precision for multiple instances or irregular lesions, so localization lags behind classification~\citep{Saporta2022}. Small nodules and masses remain difficult because deep feature maps are coarse~\citep{Sedai2018}. \citet{Wang2017-NIH} normalize each class heatmap from their gradient-free CAM to 0--255, threshold it at two fixed levels, and draw a box around every isolated region of each thresholded map, about one to three per heatmap in their experiments; the levels are in \cref{tab:related_work_comparison}.

Foreground-attention methods isolate pathology from background: PCAN thresholds its normalized map~\citep{Zhu2022-PCAN}, and \citet{Wang2024-BAS} predict a foreground map with a hand-set per-corpus threshold. Training on annotated subsets can leak information without explicit boxes~\citep{Li2018}; the heatmap-to-box step is where these methods differ most.

\subsection{Vision transformer-based approaches for chest X-ray classification and localization}
Vision transformers (ViTs) model global dependencies directly~\citep{dosovitskiy2021-ViT} and offer another map source, but need large datasets. In non-CXR images Grad-CAM was more faithful than attention rollout~\citep{Barekatain2025}, and in a pneumothorax case study deeper networks did not necessarily improve diagnosis, while the effectiveness of Grad-CAM varied across architectures~\citep{Qiu2024}. The closest CXR ViT localizers are RGT, a hybrid with one class token~\citep{Han2023-RGT}, and LT-ViT, which uses label tokens but no box benchmark~\citep{Marikkar2023-LTViT}. In natural images, TS-CAM couples a single, class-agnostic token's attention with a class-specific CAM from the patch tokens~\citep{Gao2021-TSCAM}; MCTformer learns one token per class and reads class-specific maps from class-to-patch attention, but its output is a segmentation mask with no box rule~\citep{Xu2022-MCTFormer}. CLARiTy keeps the multi-class-token map and adds a foreground mask from region-level anatomical priors and a validation-fitted box rule. CXR box rules are otherwise fixed or set from ground-truth boxes~\citep{Li-Zhou2022}, whereas calibration on separate validation images is the established remedy~\citep{Choe2020-WSOL-Eval}; \Cref{tab:related_work_comparison} compares map source, box rule and calibration.

\begin{table*}[!t]
\centering
\caption{Comparison of weakly supervised localization methods. Thresholds are rescaled to the 0--1 normalized heatmap; other values for cited models are as published.}
\label{tab:related_work_comparison}
\begin{threeparttable}
\begingroup
\setlength{\tabcolsep}{3pt}
\renewcommand{\arraystretch}{1.0}
\footnotesize\linespread{0.9}\selectfont
\begin{tabular*}{\textwidth}{@{\extracolsep{\fill}}>{\raggedright\arraybackslash}p{0.215\linewidth}>{\raggedright\arraybackslash}p{0.275\linewidth}>{\raggedright\arraybackslash}p{0.255\linewidth}>{\raggedright\arraybackslash}p{0.145\linewidth}@{}}
\toprule
Method & Map source & Box rule (threshold) & Calibrated? \\
\midrule
\citet{Wang2017-NIH} & activations $\times$ weights & fixed 0.235 / 0.706 & no \\
Grad-CAM~\citep{Selvaraju2017-GradCAM} & gradient-weighted activations & n/a & n/a \\
PCAN~\citep{Zhu2022-PCAN} & pixel classifier $\times$ attention & fixed 0.118, map only & no \\
\mbox{ThoraX-PriorNet} \mbox{\citep{Hossain2024-ThoraX-PriorNet}} & DenseNet-121 attention & fixed 0.498 + contours & no \\
\mbox{\citet{Wang2024-BAS}} & VMamba foreground map & hand-set 0.730 NIH, 0.450 RSNA & no \\
RGT~\citep{Han2023-RGT} & single class token, hybrid & fixed 0.100 (top 10\%), class-mean box & no \\
LT-ViT~\citep{Marikkar2023-LTViT} & one token per label & n/s (qualitative) & n/a \\
MCTformer~\citep{Xu2022-MCTFormer} & class-to-patch attention & n/a (no boxes) & n/a \\
\midrule
CLARiTy (ours) & class-token attention $\times$ SegmentCAM & hysteresis gate, fitted per finding & yes \\
\bottomrule
\end{tabular*}
\begin{tablenotes}
\item n/s: not stated; n/a: not applicable. Calibrated: the box rule is fitted on validation images; otherwise it is fixed or set by hand.
\item Prior supervision: region-of-interest masks for ThoraX-PriorNet, region-level priors only for CLARiTy, a foreground map that is not anatomical for Wang~2024, none elsewhere.
\item Across corpora: ThoraX-PriorNet retrains on each target corpus, Wang~2024 is evaluated within NIH and RSNA separately, and CLARiTy keeps its weights and refits only the OneBox parameters on RSNA Pneumonia and VinDr-CXR.
\end{tablenotes}
\endgroup
\end{threeparttable}
\end{table*}

\subsection{Addressing biases, shortcut learning, and explainability in chest X-ray classification models}
CXR corpora carry bias from demographic underrepresentation and from labels extracted from reports, causing underdiagnosis~\citep{SeyyedKalantari2021,Rafferty2025}. Models exploit spurious cues and generalize poorly~\citep{DeGrave2021}; Grad-CAM revealed a CNN associating chest drains with pneumothorax~\citep{Baltruschat2019}. Anatomical priors constrain predictions to plausible regions through lung-region cropping or region-specific attention~\citep{Liu2019,Hossain2024-ThoraX-PriorNet}. These gaps motivate a localizer with class-specific tokens, background suppression and anatomical priors.

\subsection{Transfer and external evaluation}
A second gap concerns evaluation: only ThoraX-PriorNet carries an anatomical prior between NIH and VinDr-CXR, but retrains on each corpus with a fixed box threshold~\citep{Hossain2024-ThoraX-PriorNet}. Whether localization transfers with frozen weights and only the box rule refitted is open; \Cref{sec:methods} sets out the model, calibration and protocol.
\section{Methods}\label{sec:methods}
\label{methodology}

\subsection{Model architecture}
\begin{figure}[tbp]
\centering
\includegraphics[width=\linewidth]{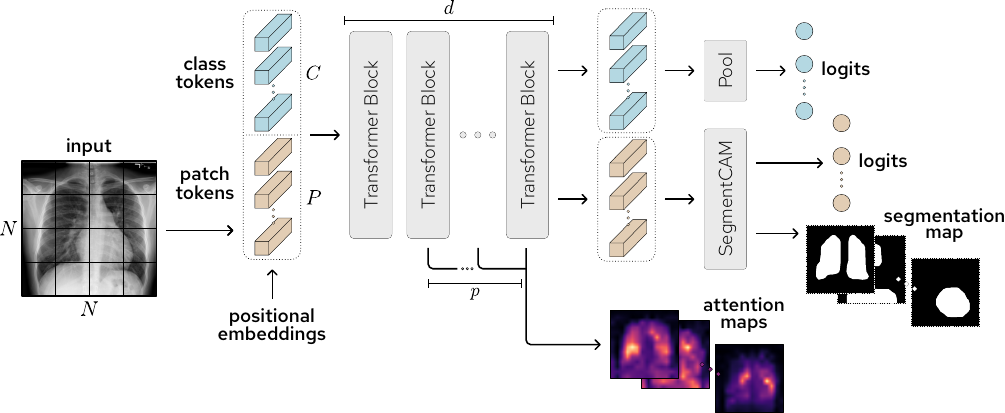}
\caption{Illustration of the proposed CLARiTy model. An input chest X-ray image is split into $N^2$ patches and embedded into $P$ patch tokens, which are concatenated with $C$ class tokens. Learned positional embeddings are added to produce the $C+P$ transformer inputs. A series of $d = 12$ transformer blocks extracts information for classification and weakly-supervised localization. At the output, the $C$ class tokens are passed through an attention pooling module to produce class token logits, and the $P$ patch tokens through the SegmentCAM module, which produces foreground logits and a segmentation map; the self-attention maps of the final $p$ blocks are fused into class-specific attention maps.}\label{fig_main-architecture}
\end{figure}

CLARiTy (\cref{fig_main-architecture}) is a ViT-S/16 backbone with $d = 12$ transformer blocks, 6 attention heads, embedding size $D = 384$ and patch size 16~\citep{dosovitskiy2021-ViT}; it partitions the $M \times M$ input into $N^2$ patches, projected to $P = N^2$ patch tokens, and uses $C$ class tokens in place of a standard transformer's single class token, one per pathology. Attention pooling maps class tokens to logits $\boldsymbol{\ell}_{C}$. A SegmentCAM module maps the patch tokens to foreground logits $\boldsymbol{\ell}_{\negthinspace f}$ and a foreground map $\boldsymbol{S}_{\negthinspace f} \in \{0,1\}^{M \times M \times C}$, while the self-attention yields class-specific attention maps $\boldsymbol{A} \in {\left( 0,1 \right)}^{N \times N \times C}$. At inference the classification score is the mean of $\boldsymbol{\ell}_{C}$ and $\boldsymbol{\ell}_{\negthinspace f}$, and boxes are extracted from the localization heatmap $\boldsymbol{\mathit{\Xi}}$ of \cref{sec:localization}.

\subsection{Weakly-supervised localization}\label{sec:localization}
Localization heatmaps combine class-specific attention maps with foreground masks.

\subsubsection{Class-specific attention maps}
The class-to-patch sub-matrix of the self-attention (\cref{fig_self-attn}) gives one attention map per class token, averaged over the final $p$ blocks:
\begin{equation}
    \boldsymbol{A}^{\left(i,j,c\right)} = \frac{1}{p} \sum_{l}^{p} \boldsymbol{A}_{l}^{\left(i,j,c\right)} , \quad \boldsymbol{A}_{l} \in {\left( 0,1 \right)}^{N \times N \times C}
\end{equation}
\noindent where $\boldsymbol{A}_{l}$ is the map at layer $l$, and $(i,j)$ and $c$ index patches and classes, respectively. We use $p = 5$ for CLARiTy-S-16-224 and $p = 3$ for CLARiTy-S-16-512, selected on the validation Unified Detection Proficiency (UDP; \cref{sec:metrics}) in the ablation study (\cref{sec:ablation-study}).

\subsubsection{Heatmaps}
Attention maps are bicubically upsampled from the $N \times N$ patch grid to $M \times M$ at inference and multiplied element-wise by the class foreground mask:
\begin{equation}
    \boldsymbol{\mathit{\Xi}}^{\left(i,j,c\right)} = \boldsymbol{A}^{\left(i,j,c\right)} \thinspace \boldsymbol{S}_{\negthinspace f}^{\,\left(i,j,c\right)}
\end{equation}
\noindent where $\boldsymbol{\mathit{\Xi}} \in \left( 0,1 \right)^{M \times M \times C}$ is the localization heatmap. The mask confines each heatmap to class-specific foreground, the anatomical region priors (\labelcref{sec:gt-seg}) add a prior over class locations, and fused heatmaps are normalized before box extraction (\cref{fig_local-heatmap}).

\subsection{Attention pooling}
The $C$ class tokens are pooled by a learned, image-invariant weighting over the $D$ embedding dimensions rather than by the average pooling used for class tokens in MCTformer~\citep{Xu2022-MCTFormer}. A weight matrix $\boldsymbol{X} \in \mathbb{R}^{C \times D}$, shared across images, is normalized by a row-wise softmax over the $D$ embedding dimensions (\labelcref{appx:architecture}):
\begin{equation}
    \mathrm{AttnPool} \left( \boldsymbol{T}, \boldsymbol{X} \right) ^ {\left(c\right)} = \sum_{\varphi=1}^{D} \boldsymbol{T}^{\left(c,\varphi\right)} \, \mathrm{softmax} \left( \boldsymbol{X} \right) ^ {\left(c,\varphi\right)} = \boldsymbol{\ell}_{C}^{\left(c\right)}
\end{equation}
\noindent where $\boldsymbol{T} \in \mathbb{R}^{C \times D}$ is the output class-token matrix. The class-token loss $\mathcal{L}^{C}_{\mathrm{CLS}}$ is the weighted binary cross-entropy (WBCE, \labelcref{appx:wbcE}) of $\boldsymbol{\ell}_{C}$ against the labels $\boldsymbol{y} \in \{0,1\}^{C}$, with sigmoid $\sigma\left(x\right) = 1 / \left(1 + \exp\left(-x\right)\right)$.

\subsection{SegmentCAM feature and heatmap extraction}
The feature stage of SegmentCAM, inspired by~\citet{Zhai2023-BAS}, reshapes the patch tokens to $N \times N \times D$ and applies two convolutional heads (\cref{fig_segcam-feat-heat}): a feature head producing $\boldsymbol{F} \in \mathbb{R}^{N \times N \times C}$, and a progressive fusion heatmap head predicting $\boldsymbol{H} \in \left(0,1\right)^{M \times M \times C}$ at input resolution. A fixed threshold of $0.5$ on $\boldsymbol{H}$ gives the foreground map $\boldsymbol{S}_{\negthinspace f} \in \{0,1\}^{M \times M \times C}$, and the background map $\boldsymbol{S}_{b} \in \{0,1\}^{N \times N \times C}$ is computed on the feature grid (\cref{sec:segcam-background}). At inference the foreground mask is therefore predicted by the network from the image alone; the region masks are training targets only (\labelcref{sec:gt-seg}).

\subsubsection{Segmentation loss}
The heatmap is trained with the soft Dice loss against the anatomical region-of-interest masks (\labelcref{sec:gt-seg}), computed per class plane at input resolution without resampling:
\begin{equation}
    \mathcal{L}_{\mathrm{Dice}} = 1 - \frac{1}{\lvert \Omega \rvert} \sum_{c \in \Omega} \frac{2 \sum_{i=1}^M \sum_{j=1}^M \boldsymbol{H}_{c}^{\left(i,j\right)} \boldsymbol{S}_{c}^{\left(i,j\right)} + \varepsilon}{\sum_{i=1}^M \sum_{j=1}^M \boldsymbol{H}_{c}^{\left(i,j\right)} + \sum_{i=1}^M \sum_{j=1}^M \boldsymbol{S}_{c}^{\left(i,j\right)} + \varepsilon} \label{eqn_L_Dice}
\end{equation}
\noindent where $c$ indexes class planes, $\boldsymbol{S}_{c}$ is the ground-truth mask, $\Omega$ is the set of planes with a non-empty mask, and $\varepsilon = 1 \times 10^{-6}$ is a constant.

\subsection{SegmentCAM background activation suppression}\label{sec:segcam-background}
The background stage (\cref{fig_segcam-back-sup}) masks $\boldsymbol{F}$ with the background map $\boldsymbol{S}_{b}$, obtained by bilinearly downsampling the heatmap to the $N \times N$ feature grid before thresholding. A shared classification head processes the masked and unmasked features, and global average pooling gives the foreground, patch-token and background logits $\boldsymbol{\ell}_{\negthinspace f}$, $\boldsymbol{\ell}_{\hspace{-0.08em} P}$ and $\boldsymbol{\ell}_{b}$ (\labelcref{appx:architecture}). The foreground and patch-token losses are WBCE terms: $\mathcal{L}^{f}_{\mathrm{CLS}} = \mathcal{L}_{\mathrm{CLS}}(\boldsymbol{\ell}_{\negthinspace f}, \boldsymbol{y})$ and $\mathcal{L}^{P}_{\mathrm{CLS}} = \mathcal{L}_{\mathrm{CLS}}(\boldsymbol{\ell}_{\hspace{-0.08em} P}, \boldsymbol{y})$.

\subsubsection{Background activation suppression}\label{sec:bas}
The BAS loss reduces the background logits relative to the patch-token logits. We adapt the single-label formulation of~\citet{Zhai2023-BAS} to multilabel classification by averaging the ratio over all positive classes, with softplus in place of ReLU (\labelcref{appx:architecture}):
\begin{equation}
    \mathcal{L}_{\mathrm{BAS}} = \frac{1}{\max\left(\sum_{c=1}^C \boldsymbol{y}^{\left(c\right)}, 1\right)} \sum_{c=1}^C \boldsymbol{y}^{\left(c\right)} \frac{ \mathrm{softplus}\left( \boldsymbol{\ell}_{b}^{\left(c\right)} \right) }{ \mathrm{softplus}\left( \boldsymbol{\ell}_{\hspace{-0.08em} P}^{\left(c\right)} \right)  + \varepsilon}
\end{equation}
\noindent where $\boldsymbol{\ell}_{b}^{\left(c\right)}$ and $\boldsymbol{\ell}_{\hspace{-0.08em} P}^{\left(c\right)}$ are the background and patch-token logits of class $c$.

\subsection{Combined loss}
The loss functions are combined with scalar weights $\alpha_C$, $\alpha_{\hspace{-0.08em} P}$, $\alpha_{\negthinspace f}$, $\beta$ and $\gamma$, tuned on validation UDP (\labelcref{appx:tuning}):
\begin{equation}
\ifarxivbuild
\begin{split}
    \mathcal{L} &= \alpha_C \mathcal{L}_{\mathrm{CLS}}^{C} + \mathcal{L}_{\mathrm{CAM}}, \\
    \mathcal{L}_{\mathrm{CAM}} &= \alpha_{\hspace{-0.08em} P} \mathcal{L}_{\mathrm{CLS}}^{P} + \alpha_{\negthinspace f} \mathcal{L}_{\mathrm{CLS}}^{f} + \beta \mathcal{L}_{\mathrm{BAS}} + \gamma \mathcal{L}_{\mathrm{Dice}}.
\end{split}
\fi
\end{equation}

\subsection{Dataset}
We use the official NIH ChestX-ray14 split~\citep{Wang2017-NIH}: 112,120 frontal radiographs of $1024 \times 1024$ pixels from 30,805 patients with 14 co-occurring pathologies, divided into a training set of 86,523 images and a test set of 25,595 images, after two blank images are excluded. Patient-wise training and validation splits (nominally 70\% and 10\%; counts in \labelcref{appx:splits}) were drawn from the training set. A subset of the test set is annotated with bounding boxes, which denote radiologist-identified pathology regions. Its 880 images carry 984 boxes for 8 of the 14 classes and were divided 50\%--50\% patient-wise into localization validation and test splits. The boxes are used only to calibrate and evaluate localization, never for training. Splits are patient-disjoint (\labelcref{appx:splits}). The 442 localization-validation images select the loss weights, checkpoint, fusion depth and OneBox parameters but never train model weights; they are part of the 25,595-image official test split on which classification AUC is reported (1.7\% of it).

\subsection{Metrics}\label{sec:metrics}
Classification is scored by the area under the receiver operating characteristic curve (AUC), macro-averaged over the 14 pathologies. For localization, a predicted box succeeds if its intersection over union (IoU) with a ground-truth box is at least $\mathrm{T(IoU)} \in \left(0,1\right)$. Macro IoU accuracy (box-level) at a given T(IoU) is the proportion of ground-truth boxes successfully localized, averaged over the eight box-annotated pathologies. On the transfer corpora, image-level IoU accuracy is the fraction of a finding's positive test images with a predicted box at IoU $\geq \mathrm{T(IoU)}$, averaged over findings. Neither version penalizes surplus boxes; \cref{appx:onebox-robustness} reports the number of boxes each extractor emits. Per-threshold results are reported at $\mathrm{T(IoU)} \in \{0.1, \allowbreak 0.2, \allowbreak 0.3, \allowbreak 0.4, \allowbreak 0.5\}$. The Multi-Scale Localization Index (MSLI) averages macro IoU accuracy over $\mathrm{T(IoU)} \in \{0.1, \allowbreak 0.25, \allowbreak 0.5, \allowbreak 0.75, \allowbreak 0.9\}$ and is the objective maximized when calibrating the box extraction. On NIH, localization is scored on all test images with boxes, regardless of classification. Intervals are 95\% stratified percentile bootstraps over 2,000 image-level resamples (\labelcref{appx:bootstrap}).

Because classification and localization must be maximized jointly, UDP is the average of macro AUC and MSLI, the objective optimized in the ablation experiments.

Model performance differs with dataset split (\cref{sec:related-work}). Reproduced on the official split, the highest reported NIH macro AUC (0.853, by MLRFNet~\citep{Li2023-MLRFNet}) shows an absolute drop of 0.052 (\cref{tab:classification_comparison}), and five models with unverified splits, reproduced for classification only, drop by 0.031 to 0.063 (\labelcref{appx:reproductions}). Results are presented together, with verified models grouped.

\subsection{Bounding boxes}\label{sec:onebox}
The ChestX-ray14 benchmark thresholds each class heatmap at two fixed levels, $0.235$ and $0.706$ on a $0$--$1$ scale ($60$ and $180$ on a $0$--$255$ scale), and draws a box around every isolated region of each thresholded map~\citep{Wang2017-NIH}. These thresholds are not calibrated to the model or data, although accuracy is sensitive to them, and the operating threshold is ill-posed without supervised data~\citep{Choe2020-WSOL-Eval}. We therefore calibrate them on the annotated localization validation split.

We use OneBox, a hysteresis-gated extractor with one box per detected finding. It (i)~optionally applies a Gaussian blur of width $\sigma$, (ii)~applies two thresholds $t_{\mathrm{lo}} < t_{\mathrm{hi}}$ for support and core regions, (iii)~keeps a support region only if it contains a core of at least four pixels (a fixed gate), and (iv)~emits one box per kept region by interpolating between its core and support boxes and scaling isotropically about its center. The high threshold locates a finding's peaks, the low threshold bounds its extent, and the gate discards weak, isolated activations. Interpolation trades precision (the core box) against coverage (the support box).

Five OneBox parameters are fitted per finding and resolution on the localization validation split: $\sigma$, $t_{\mathrm{lo}}$, $t_{\mathrm{hi}}$, the interpolation fraction $s$ between the core and support boxes and the scale $\alpha$. We evaluate $15{,}390$ configurations per finding (\labelcref{appx:onebox-space}), select the one that maximizes MSLI with ties broken in favor of fewer boxes, and freeze that setting for the test split (\cref{tab:onebox_params}). Resampling stability is reported in \labelcref{appx:onebox-robustness}.

\subsection{Training}\label{sec:training}
We distill from ConvNeXtV2-B teachers~\citep{Woo2023-ConvNeXtV2} at 224 and 512 pixels, with validation macro AUCs of $0.838$ and $0.852$ (\labelcref{appx:teacher}). To reduce training time, we use the TinyViT method~\citep{Wu2022-TinyViT} with precomputed teacher logits, but unlike the original method's sparse logits we store dense logits for every augmented image and epoch.

Images were downsampled to $224 \times 224$ or $512 \times 512$ and augmented as described in \labelcref{appx:impl-details}. CLARiTy-S-16-224 was initialized from DINO weights~\citep{Caron2021-DINO} and trained for $200$ epochs, with checkpoints selected by validation UDP; CLARiTy-S-16-512 was initialized from the trained 224 weights with interpolated positional embeddings instead of DINO pretraining.

\subsection{Transfer datasets and protocol}\label{sec:transfer-methods}

RSNA Pneumonia provides a different set of box annotations~\citep{Shih2019-RSNA} for images drawn from NIH ChestX-ray14~\citep{Wang2017-NIH}. RSNA's released test images are unlabeled, so a stratified 70/10/20 partition of its labeled images would be the natural choice for a test set. However, 72.6\% (3,875) of the 5,338 test images that partition produces map to NIH training or validation images or patients, so it would not be an external test. We therefore use one test set, ``RSNA (NIH-test-derived)'': the 7,186 RSNA images whose NIH source lies in the official ChestX-ray14 test split (\cref{tab:transfer_datasets}), with 36.0\% prevalence. Training and validation images come from the NIH-seen remainder.

\begin{table*}[tbp]
\centering
\caption{Datasets and splits. NIH (14 labels, 8 with boxes) is the primary testbed; RSNA is a transfer corpus of NIH-source images (NIH-test-derived), and VinDr-CXR is the external transfer corpus. Target train splits are used only by the RAD-DINO probe. NIH counts follow \labelcref{appx:splits}; its 493 boxes are the localization-test instances (438 images). VinDr-CXR positives span 11 labels; Edema has none in test, so macro AUC covers 10. VinDr-CXR's released 18,000 images give 17,999 here: one training image is pixel-identical to a test image and was removed.}
\label{tab:transfer_datasets}
\setlength{\tabcolsep}{1.8pt}
\begin{tabular*}{\linewidth}{@{\extracolsep{\fill}}@{}>{\raggedright\arraybackslash}p{3.0cm}>{\raggedright\arraybackslash}p{4.0cm}rrr@{}}
\toprule
Corpus & \makecell[l]{Images\\(train / val / test)} & \multicolumn{3}{c}{Test labels} \\
\cmidrule(lr){3-5}
 & & Positives & Findings & GT boxes \\
\midrule
NIH ChestX-ray14 & 75,697 / 10,826 / 25,595 & n/a & 8 & 493 \\
RSNA (NIH-test-derived) & 17,060 / 2,438 / 7,186 & 2,589 & 1 & 4,229 \\
VinDr-CXR & 13,499 / 1,500 / 3,000 & 1,243 & 6 & 932 \\
\bottomrule
\end{tabular*}
\end{table*}

VinDr-CXR~\citep{Nguyen2022-VinDrCXR} is a different corpus from another country and different hospitals, with its own annotation protocol. Its splits are image-disjoint, not patient-disjoint (no patient identifiers). Six box findings are localized (\cref{tab:transfer_vindr_per_finding}); VinDr-CXR has a single combined nodule/mass class, whereas NIH has separate Mass and Nodule classes, so that finding is scored with the larger of the NIH Mass and Nodule outputs. Five further labels (consolidation, emphysema, edema, pleural thickening, pulmonary fibrosis) are classification-only, with no NIH box annotations. Edema has no positive test image, so macro AUC covers 10 of 11 mapped labels. Image-level pneumonia and the 11 VinDr-CXR box classes with no NIH counterpart are never scored. The pleural-effusion prior adds Facies Diaphragmatica, which is the only change to the region priors.

We compare two calibration settings. The \emph{NIH-calibrated} setting keeps the NIH weights and the OneBox parameters fitted on NIH (no target data). The \emph{target-calibrated} setting keeps the same weights and heatmaps and refits the five OneBox parameters per finding on target validation boxes (2,438 RSNA and 1,500 VinDr-CXR images). The RSNA validation images come from the NIH-seen remainder, so the frozen weights saw their NIH sources in training or validation. No model weights are retrained, so classification performance is unaffected.

Neither external model's weights are trained on RSNA or VinDr-CXR box annotations, while RAD-DINO's probe is fitted on target image labels. RAD-DINO 518~\citep{PerezGarcia2025-RADDINO} freezes its encoder and fits a supervised logistic probe per finding on target training labels, with regularization and HiResCAM block chosen on validation. TorchXRayVision DenseNet121-NIH 224~\citep{Cohen2022-TorchXrayVision} applies its native head without classifier fitting; its HiResCAM layer is chosen on validation. Each is read through its own OneBox extractor, with parameters fitted on the target validation images as in the target-calibrated setting, on gradient-based HiResCAM maps~\citep{Draelos2020-HiResCAM} (used rather than Grad-CAM, whose spatially averaged gradients can highlight regions the model did not use).

Localization uses image-level IoU accuracy at $\mathrm{T(IoU)} \in \{0.1, \allowbreak 0.2, \allowbreak 0.3, \allowbreak 0.4, \allowbreak 0.5\}$ (\cref{sec:metrics}) on the transfer test splits. Intervals are 95\% percentile bootstraps ($B = 2{,}000$, seed 0, stratified image-level resampling).
\section{Results}\label{sec:results}
This section reports NIH test-set results and ablations. \Cref{tab:combined_performance} compares the models. Per-class results, computing resources, qualitative outputs (\cref{fig:clarity-localization-outputs}) and ablation details are in \labelcref{appx:results-extra,appx:ablation}.
\subsection{CLARiTy-S-16-224}
At matched resolution, CLARiTy's macro IoU accuracy over the eight findings is above every compared method at every reported threshold (\cref{tab:combined_performance}). At $\text{T(IoU)} = 0.5$, macro IoU accuracy is 0.303 (95\% confidence interval 0.271--0.333) at 224 px and 0.295 (0.263--0.331) at 512 px, against 0.201 and 0.217 published for ThoraX-PriorNet, a relative gain of 50.7\% at 224 px. Those values are published point estimates on the 880-image NIH box set, of which our 438-image, 493-box test split is one half, so no interval or paired test is possible. At matched resolution the ThoraX-PriorNet values lie below our lower 95\% limit in nine of the ten resolution--threshold cells (the exception is 512 px at T(IoU)=0.1, with 0.803 inside [0.789, 0.854]), and every other compared published value does so at its matched resolution (the 1024 px model of \citet{Wang2017-NIH} lies below both limits). At $\text{T(IoU)} = 0.5$, the 224 px model reaches 0.154 for nodules, where no compared method exceeds 0.013, and 0.326 for masses against 0.294 for the best compared 512 px model (\cref{tab:localization_comparison}). These are the two findings with the smallest median box areas (0.5\% and 1.6\% of the image), and the 224 px model beats every compared 512 px model on both from $\text{T(IoU)} = 0.2$, losing only at 0.1. The advantage is over the published models, since our 512 px model is better on nodules (0.231). Scored with the uncapped fixed-threshold rule of \citet{Wang2017-NIH} instead of OneBox, the same heatmaps exceed every published value at every threshold, at 2.4 to 3.5 times the box count (\cref{appx:onebox-robustness}). CLARiTy-S-16-224 has a macro AUC of 0.808 (0.804--0.812), above all five official-split models we reproduced (0.753--0.806), and 12 of the 23 official-split models in \cref{tab:classification_comparison} report a higher value (6 a value above the 512 px model's 0.825).

\begin{table*}[htbp]
\centering
\caption{Classification and weakly-supervised localization performance of different models evaluated on the NIH test set. Classification results are macro averaged across all 14 pathologies, and the localization results are macro averaged across the 8 pathologies with bounding box labels. Localization results for the CLARiTy models are obtained with the calibrated OneBox extractor. The highest metric values are shown in bold. The results of cited models are taken from their publications.}
\label{tab:combined_performance}
\begin{threeparttable}
\footnotesize
\setlength{\tabcolsep}{0.2pt}
\begin{tabular*}{\textwidth}{@{\extracolsep{\fill}} l c c S[table-format=1.3,detect-weight=true,table-space-text-post={\textsuperscript{\textdagger}}] *{5}{c}}
\toprule
{Model} & {Size} & \makecell{NIH\\train} & {AUC} & \multicolumn{5}{c}{\makecell{macro IoU accuracy\\at T(IoU)}} \\
\cmidrule{5-9}
& & & & {0.1} & {0.2} & {0.3} & {0.4} & {0.5} \\
\midrule
RGT\textsuperscript{1}\textsuperscript{\ddag} & 224 & 70\% & 0.839\textsuperscript{\textdagger} & 0.591 & 0.424 & 0.283 & 0.173 & 0.090 \\
ThoraX-PriorNet-224\textsuperscript{2}\textsuperscript{\ddag} & 224 & 70\% & 0.844 & 0.666 & 0.509 & 0.398 & 0.296 & 0.201 \\
ThoraX-PriorNet-512\textsuperscript{2}\textsuperscript{\ddag} & 512 & 70\% & \bfseries 0.847 & 0.803 & 0.626 & 0.488 & 0.334 & 0.217 \\
\citet{Wang2024-BAS}\textsuperscript{\ddag} & 224 & \text{--} & \text{--} & 0.450 & \text{--} & 0.230 & \text{--} & \text{--} \\
\midrule
\citet{Wang2017-NIH} & 1024 & 70\% & 0.745 & 0.568 & 0.373 & 0.221 & 0.116 & 0.062 \\
\citet{Li-Zhou2022} & 512 & 80\% & 0.829 & 0.639 & 0.527 & 0.397 & 0.314 & 0.243 \\
PCAN\textsuperscript{3} & 512 & 80\% & 0.830 & 0.778 & 0.574 & 0.364 & 0.207 & 0.103 \\
\midrule
CLARiTy-S-16-224 & 224 & 70\% & 0.808 & 0.780 & 0.667 & 0.529 & 0.401 & \bfseries 0.303 \\
CLARiTy-S-16-512 & 512 & 70\% & 0.825 & \bfseries 0.822 & \bfseries 0.698 & \bfseries 0.552 & \bfseries 0.419 & 0.295 \\
\bottomrule
\end{tabular*}
\begin{tablenotes}
\item NIH train = share of the official NIH training split used by the cited model.
\item \textsuperscript{\textdagger} Classification performance evaluated on 8 pathologies: Atelectasis, Cardiomegaly, Effusion, Infiltration, Mass, Nodule, Pneumonia, and Pneumothorax.
\item \textsuperscript{\ddag} We could not verify that published model results used the official NIH dataset split.
\item \textsuperscript{1} \citet{Han2023-RGT} \quad \textsuperscript{2} \citet{Hossain2024-ThoraX-PriorNet} \quad \textsuperscript{3} \citet{Zhu2022-PCAN}
\end{tablenotes}
\end{threeparttable}
\end{table*}

\subsection{CLARiTy-S-16-512}
Against the 224 px variant, the 512 px variant is higher in point estimate of macro IoU accuracy at $\text{T(IoU)} = 0.1$--0.4 and level at 0.5 (0.295 against 0.303), with all intervals overlapping. Its macro AUC of 0.825 (95\% confidence interval 0.821--0.829) is 0.017 above the 224 variant, so classification is higher at 512 px, although the two variants also differ in initialization and in $p$ (\cref{sec:discussion}).

\subsection{Cross-dataset transfer}\label{sec:transfer-results}
With the NIH weights frozen and no weight training on the target, CLARiTy localizes findings on RSNA Pneumonia~\citep{Shih2019-RSNA} (NIH-test-derived split) and VinDr-CXR~\citep{Nguyen2022-VinDrCXR}.

\begin{figure}[htbp]
\centering
\includegraphics[width=\linewidth]{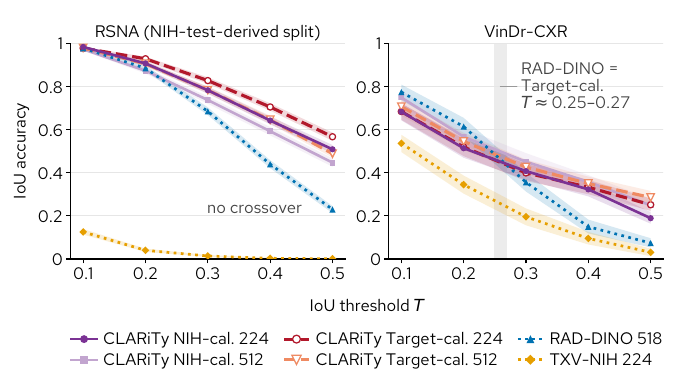}
\caption{Image-level IoU accuracy (fraction of a finding's positive test images with a box at IoU $\geq T$) on RSNA (NIH-test-derived split, 7,186 images) and VinDr-CXR (3,000 images; mean over 6 findings). CLARiTy curves are solid (NIH-calibrated, filled markers) or dashed (target-calibrated, open markers), dark 224, light 512; the two external models are dotted. Bands: 95\% bootstrap CIs. The CLARiTy NIH-cal. 224 and Target-cal. 512 curves coincide within 0.005 on RSNA and the two CLARiTy 224 px curves within 0.01 on VinDr-CXR, in both cases up to $T=0.4$.}
\label{fig:transfer-iou-vs-T}
\end{figure}

Refitting the OneBox parameters on target validation boxes changes accuracy as follows. Image-level IoU accuracy at $T(\mathrm{IoU}) = 0.1$--$0.5$ is in \cref{fig:transfer-iou-vs-T} and \cref{tab:transfer_per_T_extra}. The refit raises accuracy at $T=0.5$ in all four cases, from 0.1889 to 0.2514 on VinDr-CXR at 224 px and from 0.4469 to 0.4886 on RSNA at 512 px. On VinDr-CXR at 512 px it trades loose-threshold for strict-threshold accuracy, with 0.7491 falling to 0.7068 at $T=0.1$, $T=0.2$--$0.4$ staying within 0.03 of the NIH-calibrated value, and 0.2484 rising to 0.2851 at $T=0.5$. On VinDr-CXR at 224 px no threshold below 0.5 moves by more than 0.010. On RSNA it is level at $T=0.1$ and higher from $T=0.2$.

Against the two external models the ordering depends on the threshold. On RSNA the target-calibrated CLARiTy is level with RAD-DINO at $T=0.1$ (both 0.978) and ahead from $T=0.2$ (0.929 against 0.885 at 224 px; 0.905 against 0.885 at 512 px). On VinDr-CXR RAD-DINO leads at $T=0.1$--$0.2$ by 0.07--0.10, the curves cross near IoU 0.25--0.27 (interpolated from thresholds 0.1 apart), and CLARiTy is ahead from $T=0.3$ (0.426 against 0.356 at 512 px). TorchXRayVision DenseNet121-NIH (TXV-NIH) is behind at every threshold.

RAD-DINO was trained on all of the NIH official test list, and TXV-NIH publishes no training list and was trained on a split of the whole NIH corpus, so it may have seen part of that list as well. Both may therefore have seen the RSNA evaluation images, so their RSNA numbers are probably optimistic. VinDr-CXR, a different corpus, is not part of either model's documented pretraining data.

\Cref{tab:transfer_published} adds published numbers of in-domain methods, each under its published box convention: one enclosing box per image on RSNA, as in \citet{Wang2024-BAS}, and every contour with a strict inequality on VinDr-CXR, as in \citet{Hossain2024-ThoraX-PriorNet}. Every CLARiTy configuration, including the NIH-calibrated one, is above the published value at every threshold the published authors report, although those values come from in-domain trained models scored under a different protocol, so they are context rather than a protocol-matched comparison. At $T=0.5$ on VinDr-CXR, for example, the target-calibrated 512 model reaches 0.2851 against 0.0598. Per-finding results are in \cref{tab:transfer_vindr_per_finding} and qualitative overlays in \cref{fig:transfer-overlays}, both in \labelcref{appx:transfer}.

Classification is unchanged by recalibration, since the two calibrations share weights: macro AUC is 0.7652 (95\% CI 0.7539--0.7771) at 224 px and 0.7744 (0.7635--0.7862) at 512 px on RSNA, and 0.8826 (0.8754--0.8899) and 0.9022 (0.8952--0.9093) on VinDr-CXR.

\begin{table*}[tbp]
\centering
\caption{Published box-localization numbers beside the CLARiTy settings, each corpus under the box convention of its published comparison. RSNA: one enclosing box per image (the rule of \citet{Wang2024-BAS}). VinDr-CXR: every contour with a strict inequality (the rule of \citet{Hossain2024-ThoraX-PriorNet}). Published rows: reported by the authors, in-domain trained, protocol not identical. Entries are image-level IoU accuracy at $T$, macro-averaged over findings. Dashes mark thresholds the authors do not report. The four CLARiTy rows are the settings of \cref{tab:transfer_per_T_extra}, rescored here under this corpus's box convention; the RSNA values are therefore not the per-threshold values of that table. The highest value for each corpus and threshold is shown in bold.}
\label{tab:transfer_published}
\setlength{\tabcolsep}{3pt}
\footnotesize
\begin{tabular}{>{\raggedright\arraybackslash}p{5.1cm} c c c c c}
\toprule
 & \multicolumn{5}{c}{$T$ (IoU threshold)} \\
\cmidrule(lr){2-6}
Method & 0.1 & 0.2 & 0.3 & 0.4 & 0.5 \\
\midrule
\multicolumn{6}{l}{\textbf{RSNA (NIH-test-derived split)}} \\
\citet{Wang2024-BAS}, reported & 0.44 & --- & 0.19 & --- & --- \\
CLARiTy NIH-cal., 224 & 0.8675 & --- & 0.3357 & --- & --- \\
CLARiTy NIH-cal., 512 & 0.8355 & --- & 0.2796 & --- & --- \\
CLARiTy Target-cal., 224 & \bfseries 0.9054 & --- & \bfseries 0.4365 & --- & --- \\
CLARiTy Target-cal., 512 & 0.8687 & --- & 0.3461 & --- & --- \\
\midrule
\multicolumn{6}{l}{\textbf{VinDr-CXR (official test split)}} \\
\citet{Hossain2024-ThoraX-PriorNet}, VinDr priors, reported & 0.4565 & 0.3536 & 0.2468 & 0.1446 & 0.0598 \\
CLARiTy NIH-cal., 224 & 0.6832 & 0.5138 & 0.4073 & 0.3226 & 0.1889 \\
CLARiTy NIH-cal., 512 & \bfseries 0.7491 & \bfseries 0.5671 & \bfseries 0.4511 & \bfseries 0.3576 & 0.2484 \\
CLARiTy Target-cal., 224 & 0.6840 & 0.5191 & 0.3992 & 0.3326 & 0.2514 \\
CLARiTy Target-cal., 512 & 0.7068 & 0.5454 & 0.4261 & 0.3495 & \bfseries 0.2851 \\
\bottomrule
\end{tabular}
\end{table*}

\subsection{Ablation study}\label{sec:ablation-study}
\begin{figure}[htbp]
\centering
\includegraphics[width=\linewidth]{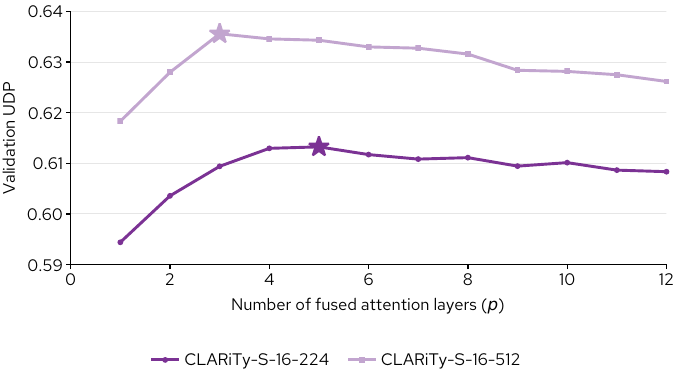}
\caption{Validation UDP against the number of fused attention layers $p$ for CLARiTy-S-16-224 and CLARiTy-S-16-512. Each point is that variant's highest validation UDP during training, so the curves are selected on validation data and small differences between neighboring values of $p$ are not a ranking. Stars mark the curve maxima.}
\label{fig:p-max-val-UDP}
\end{figure}

All ablations except the attention-layer sweep use CLARiTy-S-16-224 on NIH with one seed (\cref{tab:ablation_compact}). Removing SegmentCAM lowers validation MSLI by a relative 13.3\%, and removing its Dice loss by a relative 22.8\%, the largest localization drop. Without SegmentCAM, OneBox reads the class-token attention alone. Without the Dice loss, the foreground mask is trained only by the classification and background-suppression losses, and MSLI falls to 0.302, below the 0.339 of the variant without SegmentCAM (one seed). That gap suggests that a mask trained without the anatomical Dice term does worse than no mask. ImageNet instead of DINO weights, global average pooling (GAP) instead of attention pooling, and the multi-label soft-margin loss (MLSM) instead of WBCE lower MSLI by a relative 7.0\%, 6.4\% and 2.3\%. GAP raises validation macro AUC from 0.836 to 0.843. Fusing the attention maps of a subset of the final layers gives the most class-discriminative heatmaps, with validation UDP peaking at $p=5$ (224 px) and $p=3$ (512 px) (\cref{fig:p-max-val-UDP}).

\begin{table}[htbp]
\centering
\caption{Component ablations on CLARiTy-S-16-224, with one component changed per row relative to the baseline. Metrics are on the validation set, with localization from the calibrated OneBox extractor. MSLI is the macro IoU accuracy averaged over T(IoU) $\in \{0.1, 0.25, 0.5, 0.75, 0.9\}$, and UDP is the mean of macro AUC and MSLI. The highest values are shown in bold. Relative changes quoted in the text use unrounded values. Scores are validation values on the split used to fit the OneBox parameters and $p$, from one seed, so they rank the components rather than estimate their size.}
\label{tab:ablation_compact}
\resizebox{\columnwidth}{!}{%
\begin{tabular}{l c c c}
\toprule
Setting & macro AUC & MSLI & UDP \\
\midrule
Baseline (all components) & 0.836 & \bfseries 0.391 & \bfseries 0.614 \\
SegmentCAM off (and its Dice loss) & 0.841 & 0.339 & 0.590 \\
Dice loss off & 0.833 & 0.302 & 0.567 \\
ImageNet weights (not DINO) & 0.826 & 0.363 & 0.595 \\
MLSM loss (not WBCE) & 0.841 & 0.382 & 0.611 \\
GAP pooling (not attention) & \bfseries 0.843 & 0.366 & 0.604 \\
\bottomrule
\end{tabular}}
\end{table}

\section{Discussion}
\label{sec:discussion}
At matched resolution (\cref{sec:results}), CLARiTy's lead over the strongest published localizer holds at every reported threshold and widens as the criterion tightens, reaching a 50.7\% relative gain at T(IoU)=0.5 against the published ThoraX-PriorNet-224 value. The largest relative gain is on nodules, where the 224 px model reaches 0.154 at IoU 0.5 against at most 0.013 for any compared method. On masses it is 2.8 times ThoraX-PriorNet-224 at matched resolution (0.326 against 0.118), but only 11\% above the best compared method, ThoraX-PriorNet-512 (0.294). Coarse feature maps and small box areas make these two findings the hardest to localize (\cref{appx:failures}).

The anatomical prior contributes, but it does not carry the result. It enters only as the training target of the SegmentCAM mask, and six of the eight evaluated findings share the same lung prior; averaged over the boxed localization images, the region priors overlap the ground-truth boxes at a mean IoU of only 0.19 (\labelcref{sec:gt-seg}). Removing SegmentCAM, and with it every use of the prior, lowers validation MSLI by 13.3\% (0.339 against 0.391), whereas removing only its Dice supervision lowers it by 22.8\%, which suggests that a mask trained without the anatomical Dice term does worse than no mask. Cardiomegaly, whose prior is the heart, is scored 1.000 at every threshold by both CLARiTy models; without it, the 224 px model keeps its lead over ThoraX-PriorNet-224 at every threshold, and the 512 px model keeps its lead up to IoU 0.4 and is level with ThoraX-PriorNet-512 at IoU 0.5 (0.194 against 0.192).

The uncapped fixed-threshold rule of \citet{Wang2017-NIH}, applied to the same heatmaps, exceeds every published value at every threshold, but only by emitting 2.4 to 3.5 times as many boxes (\cref{appx:onebox-robustness}). At one box per instance, at 224 px, it falls below the published ThoraX-PriorNet-224 values (0.144 against 0.201 at IoU 0.5), whereas OneBox reaches 0.261, so the box rule matters at equal budget. OneBox gives a compact output, and refitting its parameters on target validation images, with the weights frozen, carries the extractor to a new corpus and raises accuracy where boxes must fit tightly. Classification is competitive, not leading: the NIH AUCs (0.808 and 0.825) exceed all five official-split models we reproduced, but 12 and 6 of the 23 official-split models in \cref{tab:classification_comparison} report higher values.

Those gains are conditional, as the ordering against the external models depends on the threshold: on VinDr-CXR the curves cross near IoU 0.25--0.27 (interpolated from thresholds 0.1 apart), with RAD-DINO ahead at looser thresholds and CLARiTy ahead from IoU 0.3. Resolution is not decisive: the 512 model leads at loose thresholds and on nodules at every threshold, and the 224 model is nominally ahead at the strictest threshold, where it is also ahead on masses (0.326 against 0.302 for CLARiTy-S-16-512). The 512 model is initialized from the 224 weights and fuses $p=3$ rather than $5$ layers, so the pair is not a resolution-only comparison.

The RSNA comparison is not independent: both external models may have seen RSNA images during pretraining, so their RSNA scores probably flatter them. VinDr-CXR is not part of either model's documented pretraining data.

RSNA is a second annotation of the same images: it supplies new boxes for ChestX-ray14, and because most of its labeled images map to NIH training or validation images, we evaluate only on the NIH-test-derived subset of 7,186 images. Our model weights never train on bounding boxes; only the OneBox parameters are fitted on target validation images. VinDr-CXR is external (another country and hospitals) with image-disjoint, not patient-disjoint, splits.

Several limitations bound these results. The ablations use one training seed and are scored on validation images, so they rank components rather than size them, and the confidence intervals capture resampling of the test images, not variation between training runs. The smallest findings are also the least certain, since nodules and masses have only 39 and 43 localization-test instances.

The comparisons are not fully symmetric. Neither a Grad-CAM variant of CLARiTy nor a reproduced baseline is run through OneBox on NIH, so on NIH the effect of OneBox is isolated only against the rule of \citet{Wang2017-NIH} on CLARiTy's own heatmaps. The transfer study does run two external models through OneBox, but only on RSNA and VinDr-CXR. The target-calibrated setting also uses information that the NIH-calibrated one does not: the OneBox refit uses target-corpus boxes, and the RAD-DINO probe uses target-corpus image labels. The published rows are point estimates without intervals, they are not protocol-matched, and they are scored on the 880-image box set, which includes our test half.

Finally, the data design leaves some overlap and limits what the statistics can say. The 442 localization-validation images, which select the model and its OneBox parameters, lie inside the 25,595-image test split used for classification AUC (1.7\% of it), whereas the 438 localization-test images select nothing. We report no multiplicity correction, so the cells are descriptive, and we studied two external corpora and report no reader study.
\section{Conclusion}
\label{sec:conclusion}
CLARiTy shows that a vision transformer can localize chest X-ray pathologies without gradient class activation mapping. On the official NIH test split it outperforms the strongest of the compared published localizers at matched resolution at every reported IoU threshold. The gain is clearest on nodules and masses, the two findings whose boxes cover the smallest fraction of the image. On nodules, the two models reach 0.154 (224 px) and 0.231 (512 px) at IoU 0.5, against at most 0.013 for any compared method. On masses, at matched 224 px resolution, the 224 px model reaches 0.326 against 0.118 for ThoraX-PriorNet-224, and it is 11\% above the best compared method, ThoraX-PriorNet-512. The ablations show the largest localization drops when SegmentCAM, and especially its Dice supervision, is removed, with attention pooling and self-supervised initialization contributing in the same direction.

What carries the result beyond one benchmark is the box step. OneBox turns a class heatmap into boxes through five parameters fitted per finding and resolution on held-out validation images and then frozen, so the conversion is explicit and repeatable. With the NIH weights frozen, refitting only those parameters on target validation improves image-level IoU accuracy at the strictest reported threshold in both corpora at both resolutions; on the NIH-test-derived RSNA split the refitted model is level with RAD-DINO (a self-supervised vision transformer whose probe is fitted on target labels) at the loosest threshold and ahead from IoU 0.2, and on VinDr-CXR it is numerically ahead from IoU 0.3.

What generalizes is the protocol: fit the extractor on validation images, freeze it before test, and report localization across thresholds.

\section*{CRediT authorship contribution statement}
\textbf{John M. Statheros:} Conceptualization, Methodology, Software, Validation, Formal Analysis, Investigation, Data Curation, Writing -- Original Draft, Visualization. \textbf{Hairong Wang:} Conceptualization, Methodology, Writing -- Review \& Editing, Supervision. \textbf{Richard Klein:} Conceptualization, Methodology, Writing -- Review \& Editing, Supervision.

\section*{Declaration of competing interest}
The authors declare that they have no known competing financial interests or personal relationships that could have appeared to influence the work reported in this paper.

\section*{Data and code availability}
The source code used to train and evaluate the models in this study is available at \url{https://github.com/jstat17/clarity-cxr}. ChestX-ray14 is available from the National Institutes of Health~\citep{Wang2017-NIH} (\url{https://nihcc.app.box.com/v/ChestXray-NIHCC}), RSNA Pneumonia from Kaggle~\citep{Shih2019-RSNA} (\url{https://www.kaggle.com/competitions/rsna-pneumonia-detection-challenge}), and VinDr-CXR from PhysioNet~\citep{Nguyen2022-VinDrCXR} (\url{https://physionet.org/content/vindr-cxr/1.0.0/}). None of the three image sets can be redistributed.

\section*{Ethics statement}
This study used the publicly available, de-identified NIH ChestX-ray14 dataset. It also used the de-identified RSNA Pneumonia Detection Challenge dataset~\citep{Shih2019-RSNA} and the de-identified VinDr-CXR dataset~\citep{Nguyen2022-VinDrCXR}, the latter obtained under PhysioNet credentialed access. No new human data were collected, and no ethics approval or informed consent was required for this secondary analysis.

\section*{Acknowledgments}
Funding for infrastructure related to this work was provided by the Canadian Institutes of Health Research (CIHR) under project grant titled ``Using locally developed computer-assisted detection to promote social justice for a population with a high burden of lung disease: A participatory equity-sensitive approach'' (RN520904 -- 506651). The funders had no role in the study design, data collection and analysis, decision to publish, or preparation of the manuscript.

\section*{Declaration of generative AI and AI-assisted technologies in the manuscript preparation process}
During the preparation of this work the authors used Claude, ChatGPT and Grok in order to improve clarity and readability of prose. After using these tools, the authors reviewed and edited the content as needed and take full responsibility for the content of the published article.

\appendix
\crefalias{section}{appendix}
\crefalias{subsection}{subappendix}
\counterwithin*{table}{section}
\counterwithin*{figure}{section}
\counterwithin*{equation}{section}
\renewcommand{\theHtable}{\Alph{section}.\arabic{table}}
\renewcommand{\theHfigure}{\Alph{section}.\arabic{figure}}
\renewcommand{\theHequation}{\Alph{section}.\arabic{equation}}
\makeatletter
\providecommand{\suppressappendixfloats}{%
  \suppressfloats[t]%
  \global\@dbltopnum\z@}
\makeatother
\enableAppendixBottomFloats
\appendixfloatbarrier
\section{Implementation details}\label{appx:impl-details}
\suppressappendixfloats

\subsection{Weighted binary cross-entropy loss}\label{appx:wbcE}
The classification loss is the weighted binary cross-entropy (WBCE):
\begin{equation}
\begin{split}
    \mathcal{L}_{\mathrm{CLS}}(\boldsymbol{\ell}, \boldsymbol{y}) = - \sum_{c=1}^C \thinspace\Bigl[ & w_{\hspace{-0.08em} P} \hspace{0.1em} \boldsymbol{y}^{\left(c\right)} \ln \sigma\left(\boldsymbol{\ell}^{\left(c\right)}\right) + \\ & w_{\hspace{-0.08em} N} \left(1 - \boldsymbol{y}^{\left(c\right)}\right) \ln \left(1 - \sigma\left(\boldsymbol{\ell}^{\left(c\right)}\right)\right) \Bigr]
\end{split}
\label{eqn_L_CLS}
\end{equation}
\noindent where $\mathcal{L}_{\mathrm{CLS}}$: classification loss, $\boldsymbol{\ell} \in \mathbb{R}^{C}$: predicted logits, $\boldsymbol{y} \in \{0,1\}^{C}$: ground-truth binary labels, $w_{\hspace{-0.08em} P}$: positive-label weight, $w_{\hspace{-0.08em} N}$: negative-label weight, $\boldsymbol{y}^{\left(c\right)}$: label at index $c$, and $\boldsymbol{\ell}^{\left(c\right)}$: predicted logit at index $c$.
The weights compensate for class imbalance, $w_{\hspace{-0.08em} P} = \left(\lvert P \rvert + \lvert N \rvert\right) / \lvert P \rvert$ and $w_{\hspace{-0.08em} N} = \left(\lvert P \rvert + \lvert N \rvert\right) / \lvert N \rvert$, where $\lvert P \rvert$ and $\lvert N \rvert$ are the number of positive and negative labels in the batch.

\subsection{Loss-weight tuning}\label{appx:tuning}
Loss-weight tuning used the CLARiTy-S-16-224 model to optimize $\alpha_C$, $\alpha_{\hspace{-0.08em} P}$, $\alpha_{\negthinspace f}$, $\beta$ and $\gamma$. We used the Tree-structured Parzen Estimator (TPE) sampler of Optuna~\citep{Akiba2019-Optuna}, with 20 trials, 10 startup trials, and 100 candidate samples per trial. Each scalar weight was searched over $\left[ 0,5 \right]$, and the search maximized the validation UDP, which combines classification and weakly-supervised localization. The selected loss weights are given in \cref{tab:loss_weights}.

\subsection{Teacher training recipe}\label{appx:teacher}
The ConvNeXtV2-B teachers~\citep{Woo2023-ConvNeXtV2} (224 px and 512 px) were trained with image augmentations and the weighted binary cross-entropy loss (\labelcref{appx:wbcE}) for $100$ epochs. The settings were a batch size of $1024$, a learning rate of $2.5 \times 10^{-5}$, the AdamW optimizer, a weight decay of $0.05$, a classifier head dropout rate of 0.1 and a stochastic depth rate of 0.1, with a cosine learning schedule and a warmup of 3 epochs. Both teachers were initialized from ImageNet weights, and the checkpoint with the best validation macro AUC after each epoch was selected.

\subsection{Data split integrity}\label{appx:splits}
We audited all dataset splits used in this work. Every required pair of splits is patient-disjoint: the official training and test splits share no patients ($28{,}008$ vs $2{,}797$ patients), nor do the classification training and validation splits ($24{,}507$ vs $3{,}501$ patients) or the localization validation and test splits ($363$ patients each, no shared patients). The localization validation and test images (442 and 438 images) are subsets of the official test split, contain no official-training image or patient, and share no images with one another. The 442 localization-validation images select the loss weights, checkpoint, fusion depth and OneBox parameters but never train model weights, and classification AUC is reported on the $25{,}595$-image official test split that contains them (1.7\% of it). Finally, an exact-duplicate MD5 scan over all $112{,}120$ images found no repeated digest, so no image is shared between any two splits. Two images are excluded because they are blank (\texttt{00007160\_002.png} from the official training list and \texttt{00010007\_121.png} from the official test list), so the training, validation and test splits contain $75{,}697$, $10{,}826$ and $25{,}595$ images ($112{,}118$ in total).

\subsection{Image resolution and augmentation}
Models were trained at an image resolution of $224 \times 224$ or $512\times 512$ by downsampling the original $1024 \times 1024$ images via the Lanczos method. Training augmentations were random rotation, random cropping, random Poisson noise, and random brightness. Augmentations were generated with a pseudo-random generator seeded uniquely for each image and epoch.

\subsection{Model specification}
CLARiTy is built on the ViT-S-16 backbone, which has 12 layers, 6 self-attention heads per layer, embedding size of 384 and a patch size of 16. Each block follows the standard ViT-S/16 design~\citep{dosovitskiy2021-ViT}, with pre-layer normalization, multi-head self-attention followed by an MLP of size 1536, and residual connections. After an ablation study (\cref{sec:ablation-study}), the number of fused attention layers was set to $p=5$ for CLARiTy-S-16-224 and $p=3$ for CLARiTy-S-16-512, selected using the validation UDP.

The CLARiTy-S-16-224 configuration takes $224\times 224$ images as input, resulting in an $N \times N = 14 \times 14$ patch grid. The progressive fusion block produces $224 \times 224$ localization heatmaps directly, so no inference-time upsampling of the heatmaps is required.

The CLARiTy-S-16-512 configuration takes $512\times 512$ images as input, resulting in a $32 \times 32$ patch grid. The progressive fusion block likewise produces $512 \times 512$ localization heatmaps directly.

\subsection{Training}
The CLARiTy-S-16-224 model was initialized from DINO pretrained weights~\citep{Caron2021-DINO} and trained for $200$ epochs with a batch size of $1024$, a learning rate of $1 \times 10^{-4}$, the AdamW optimizer, a weight decay of $0.05$, and dropout rates of 0.1 for attention pooling, the MLP and self-attention, with a stochastic depth rate of 0.1. A cosine learning schedule with a warmup of 5 epochs was used. The best-performing model was selected during training via the UDP, by evaluating on the validation set after each epoch.

The CLARiTy-S-16-512 was trained with the same methodology, but was initialized from the trained CLARiTy-S-16-224 weights with interpolated positional embeddings instead of DINO pretraining.

\subsection{Comparison reproductions}\label{appx:reproductions}
We also retrained five other models whose published results we could not verify as using the official NIH split. We trained these using the official train--test split and the same patient-wise training and validation partition as CLARiTy, and tested only their classification performance. Since the original publications do not fully report their training procedures, all five reproductions followed each model's published architecture and, where available, its reported training configuration; the remaining choices followed a common protocol: ImageNet-pretrained initialization, a fixed seed, checkpoint selection by validation macro AUC, and a single evaluation on the official test split.

\appendixfloatbarrier
\section{Architecture details}\label{appx:architecture}
\suppressappendixfloats

This appendix gives the architecture details referred to in \cref{sec:methods}.

\begin{figure}[htbp]
\centering
\includegraphics[width=0.4\linewidth]{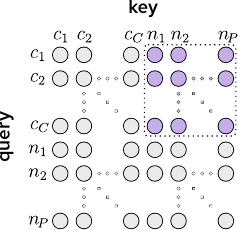}
\caption{Transformer self-attention matrix with multiple class tokens. Each class token is denoted $c_1,\allowbreak c_2,\allowbreak \dots,\allowbreak c_C$, and each patch token is denoted $n_1,\allowbreak n_2,\allowbreak \dots,\allowbreak n_{\negthinspace P}$. The class-to-patch attention in the upper-right sub-matrix is extracted for the class-specific attention maps.}\label{fig_self-attn}
\end{figure}

\begin{figure}[htbp]
\centering
\includegraphics[width=0.8\linewidth]{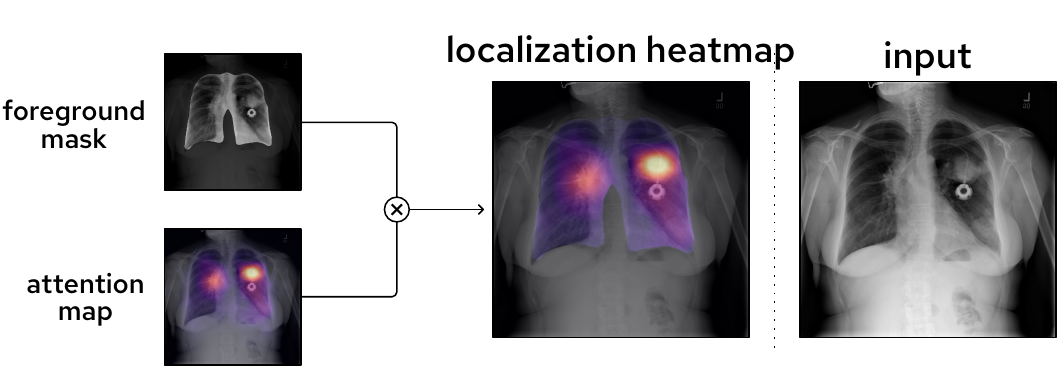}
\caption{Weakly-supervised localization method of CLARiTy. During inference, a foreground mask and attention map are produced for each class. Element-wise multiplication yields a class-specific localization heatmap confined to the class's anatomical region. In this chest X-ray, a mass (round opacity) is found in the upper-middle left lung of the patient. The attention map has high intensity directly over the mass, and the heatmap intensity is confined to the lung lobes.}
\label{fig_local-heatmap}
\end{figure}

\begin{figure}[htbp]
\centering
\includegraphics[width=\linewidth]{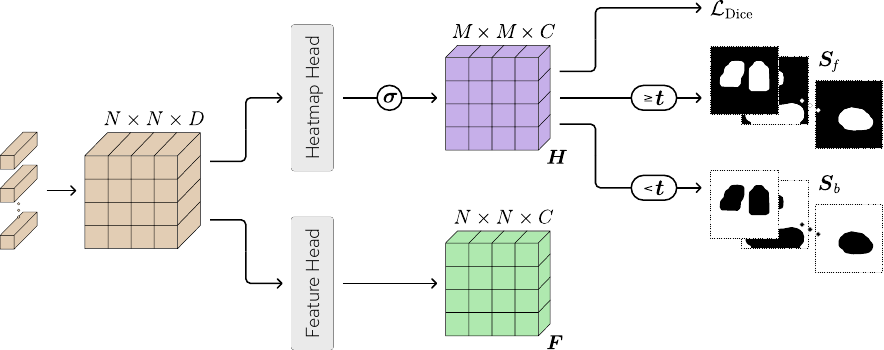}
\caption{Illustration of the SegmentCAM module's feature and heatmap extraction. The output patch tokens are reshaped to an $N \times N \times D$ tensor. Then, in the heatmap branch, the progressive fusion block (\cref{fig_prog-fusion-block}) with sigmoid activation produces a heatmap tensor $\boldsymbol{H}$ of shape $M \times M \times C$, where $M$ is the input image resolution. A threshold $t$ is applied to $\boldsymbol{H}$ to produce foreground and background segmentation maps $\boldsymbol{S}_{\negthinspace f}$ and $\boldsymbol{S}_{b}$ respectively. The heatmap $\boldsymbol{H}$ is used to calculate the soft Dice segmentation loss $\mathcal{L}_{\mathrm{Dice}}$. In the feature branch, a convolutional head produces feature tensor $\boldsymbol{F}$ of shape $N \times N \times C$, which is used in the background activation suppression methodology.}\label{fig_segcam-feat-heat}
\end{figure}

\begin{figure}[htbp]
\centering
\includegraphics[width=\linewidth]{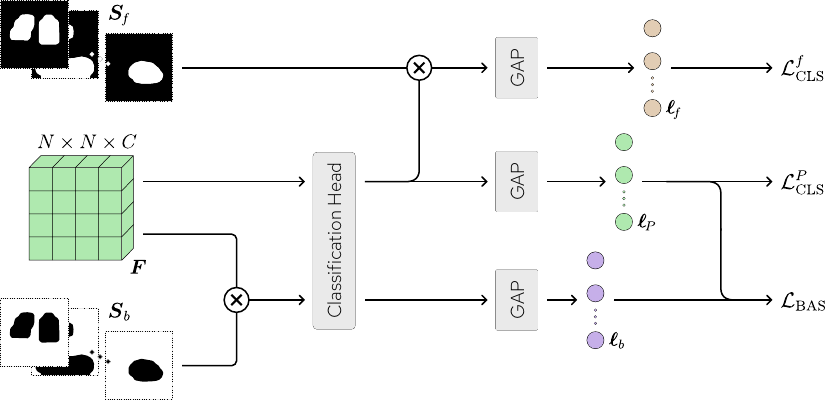}
\caption{Illustration of the SegmentCAM module's background activation suppression methodology. The segmentation maps $\boldsymbol{S}_{\negthinspace f}$ and $\boldsymbol{S}_{b}$ are derived from the heatmap at the feature resolution. The feature tensor $\boldsymbol{F}$ is passed through the convolutional classification head twice: with unmasked features, and with foreground features suppressed (after fusing with $\boldsymbol{S}_{b}$). In the top branch, background features of the unmasked output are suppressed with $\boldsymbol{S}_{\negthinspace f}$, and global average pooling (GAP) produces foreground logits $\boldsymbol{\ell}_{\negthinspace f}$. In the middle branch, the output unmasked features are used to produce patch token logits $\boldsymbol{\ell}_{\hspace{-0.08em} P}$ using GAP. In the bottom branch, the output of the foreground-suppressed features produces background logits $\boldsymbol{\ell}_{b}$ using GAP. Foreground and patch token losses $\mathcal{L}_{\mathrm{CLS}}^{f}$ and $\mathcal{L}_{\mathrm{CLS}}^{P}$ are calculated using $\boldsymbol{\ell}_{\negthinspace f}$ and $\boldsymbol{\ell}_{\hspace{-0.08em} P}$ respectively. Finally, the background activation suppression loss $\mathcal{L}_{\mathrm{BAS}}$ is calculated using $\boldsymbol{\ell}_{\hspace{-0.08em} P}$ and $\boldsymbol{\ell}_{b}$. Using these losses, the classification head learns to classify $\boldsymbol{F}$ while ignoring background features.}\label{fig_segcam-back-sup}
\end{figure}

\subsection{Class token losses and attention pooling}
\begin{figure}[htbp]
\centering
\includegraphics[width=\linewidth]{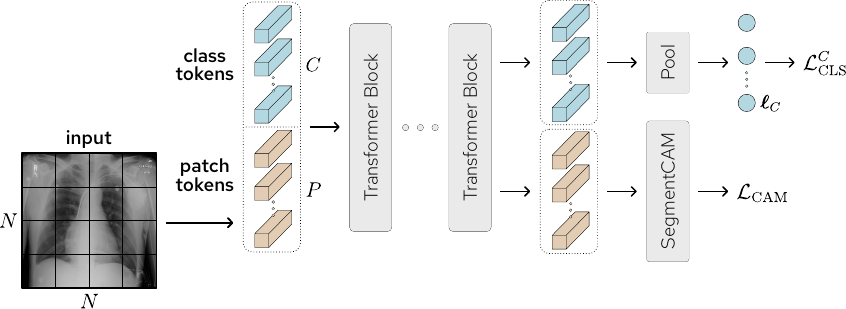}
\caption{Illustration of the CLARiTy losses. Following the attention pooling layer, the class token logits are used to calculate the class token classification loss $\mathcal{L}_{\mathrm{CLS}}^{C}$. The outputs of the SegmentCAM module are used to calculate the combined CAM loss $\mathcal{L}_{\mathrm{CAM}}$.}\label{fig_main-loss}
\end{figure}

The class token logits $\boldsymbol{\ell}_{C}$ enter the classification loss $\mathcal{L}^{C}_{\mathrm{CLS}}$, and the SegmentCAM module produces a combined CAM loss $\mathcal{L}_{\mathrm{CAM}}$. A further illustration of the proposed model is shown in \cref{fig_main-loss}.

\begin{figure}[htbp]
\centering
\includegraphics[width=0.96\linewidth]{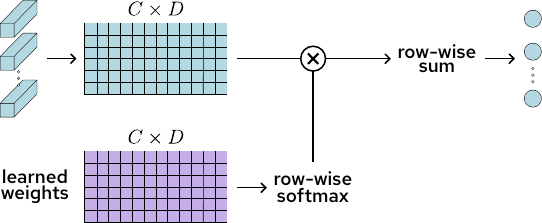}
\caption{Illustration of the attention pooling module. Row-wise softmax activation is applied to a set of learned weights, which are invariant to the input image. The resulting weight matrix is multiplied element-wise with the output class tokens, reshaped to a $C\times D$ matrix, and a row-wise sum produces the class-specific logits. Each class therefore weights the embedding dimensions independently.}\label{fig_attn-pool}
\end{figure}

Attention pooling uses a learned weight matrix $\boldsymbol{X} \in \mathbb{R}^{C \times D}$ that is shared across all input images. The $\mathrm{softmax}$ is applied row-wise over the $D$ embedding dimensions, so each class selects its own combination of embedding dimensions. \Cref{fig_attn-pool} shows the attention pooling module.

\subsection{Class-specific attention maps}
\Cref{fig_self-attn} shows how the self-attention matrix of a block is partitioned for $C$ class tokens and $P$ patch tokens. With the tokens ordered by type, the upper-left sub-matrix holds the class-to-class attention, the lower-right sub-matrix the patch-to-patch attention, and the upper-right sub-matrix the attention from each class token to each patch token. Only the class-to-patch block is extracted. For a class $c$, its row gives an $N \times N$ spatial attention map. As described in \cref{sec:methods}, this map is averaged over the final $p$ blocks, upsampled, and multiplied by the class foreground mask.

\subsection{SegmentCAM feature and heatmap extraction}
\Cref{fig_segcam-feat-heat} illustrates the two branches of the SegmentCAM feature stage: the feature tensor $\boldsymbol{F}$ and the image-resolution heatmap $\boldsymbol{H}$, which is thresholded at $t = 0.5$ to give the foreground and background maps $\boldsymbol{S}_{\negthinspace f}$ and $\boldsymbol{S}_{b}$.

The soft Dice loss is two-sided: it is minimized when $\boldsymbol{H}$ is $1$ inside the ground-truth mask and $0$ outside. Because the loss is evaluated on the raw sigmoid heatmap at image resolution, with no loss-time resampling, $\boldsymbol{H}$ stays calibrated to the threshold $t = 0.5$ applied at inference.

\subsection{Progressive heatmap head and fused inverse bottleneck}
\begin{figure}[htbp]
\centering
\includegraphics[width=0.79\linewidth]{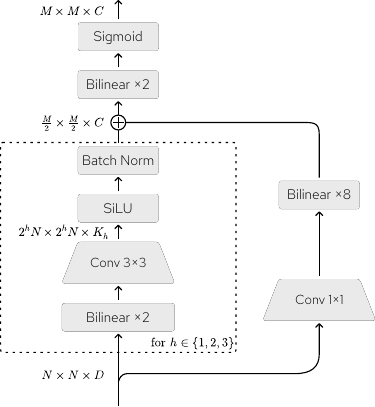}
\caption{Illustration of the progressive fusion block used in the heatmap branch of SegmentCAM. The $N \times N \times D$ patch tokens are progressively refined and upsampled by three stages of bilinear $\times 2$ interpolation, each followed by a $3 \times 3$ convolution, $\mathrm{SiLU}$ activation, and batch normalization. The channel width tapers from $\max\left(4C, 64\right)$ through $\max\left(3C, 48\right)$ to $C$, so all learned computation runs at or below half the image resolution. A $1 \times 1$ convolution projects the raw patch tokens into $C$ class maps, which are bilinearly enlarged by a factor of $8$ and added to the main path, restoring the global context that the narrowed stages would otherwise discard. A final bilinear $\times 2$ and sigmoid activation produce the image-resolution heatmap $\boldsymbol{H} \in \left(0,1\right)^{M \times M \times C}$.}\label{fig_prog-fusion-block}
\end{figure}

Both convolutional heads of the SegmentCAM feature stage reduce the channel dimension from $D$ to the number of classes $C$. The heatmap head is a progressive fusion block (\cref{fig_prog-fusion-block}) that progressively refines and upsamples the patch tokens to predict the class heatmaps directly at the input image resolution $M \times M$. It is inspired by the lateral feature fusion of feature pyramid networks~\citep{Lin2017-FPN}, with a low-resolution skip that preserves the global context of the raw patch tokens. The feature head is a fused inverse bottleneck with an expansion ratio of 2 (\cref{fig_fused-inv-bott}) and produces the feature tensor $\boldsymbol{F}$.

\subsection{Background activation suppression derivation}
\begin{figure}[htbp]
\centering
\includegraphics[width=0.372\linewidth]{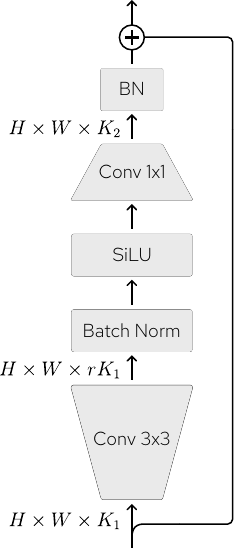}
\caption{Fused inverse bottleneck block, as used in~\citet{Xiong2021-MobileDets}. In SegmentCAM, the block is used in the feature head and in the classification head of the background activation suppression stage. An input tensor of shape $H \times W \times K_1$ is passed through a $3 \times 3$ convolution with an expansion ratio of $r \in \mathbb{R}_{> 1}$. This increases the number of channels to $rK_1$. Batch normalization and $\mathrm{SiLU}$ activation are then applied. Finally, a $1 \times 1$ convolution with batch normalization reduces the number of channels to $K_2$. If $K_1=K_2$, then a residual connection is added to the output.}\label{fig_fused-inv-bott}
\end{figure}

The suppression mask $\boldsymbol{S}_{b}$ is obtained by bilinearly downsampling the image-resolution heatmap to the $N \times N$ feature grid before thresholding. Both the suppressed and unsuppressed feature tensors are passed through the same convolutional classification head, which comprises three fused inverse bottleneck blocks, each with an expansion ratio of 2 (\cref{fig_fused-inv-bott}). After the head, the unsuppressed tensor is fused with $\boldsymbol{S}_{\negthinspace f}$ in a separate branch, and global average pooling is applied to all final feature tensors. The foreground logits $\boldsymbol{\ell}_{\negthinspace f}$ contain only foreground information, while the background logits $\boldsymbol{\ell}_{b}$ contain only background information. The patch-token logits $\boldsymbol{\ell}_{\hspace{-0.08em} P}$ contain information from the unmasked features. The background activation suppression loss $\mathcal{L}_{\mathrm{BAS}}$ is applied to both $\boldsymbol{\ell}_{\hspace{-0.08em} P}$ and $\boldsymbol{\ell}_{b}$.

The original BAS formulation targets single-label classification, where the ratio of background to overall activation is computed for the ground-truth class only~\citep{Zhai2023-BAS}. We adapt it to multilabel classification by computing the ratio for every class present in the image and averaging over them. In this setting, a present class may have a negative pooled logit, particularly early in training, so a $\mathrm{ReLU}$ activation can output exactly zero and make the ratio undefined or unstable. $\mathrm{Softplus}$, $\mathrm{softplus}\left(x\right) = \ln\left(1 + \exp\left(x\right)\right)$, is strictly positive and smooth, which keeps the ratio finite and differentiable, so we use it instead.

\subsection{Ground-truth segmentation maps}\label{sec:gt-seg}
The TorchXrayVision segmentation model produces anatomical segmentations for chest X-ray images and was used to generate the ground-truth segmentation maps~\citep{Cohen2022-TorchXrayVision}. A diagnostic radiologist identified which of the anatomical regions produced by the segmentation model each NIH pathology could be located in (\cref{tab:cxr14_regions}). The anatomical masks for each pathology are combined into a single region-of-interest mask. These region-of-interest masks are then combined into a segmentation map for each image. The region-of-interest masks are broad: averaged over the boxed localization images, they cover 26\% of the image (9--33\% per finding) and overlap the ground-truth boxes at a mean IoU of 0.19 (0.02 for nodules), so they constrain the heatmaps to plausible anatomy rather than provide localization supervision.

\begin{table}[ht]
\centering
\caption{Anatomical regions of pathologies in the NIH dataset, as identified by a diagnostic radiologist.}
\label{tab:cxr14_regions}
\begin{tabular}{l l}
\toprule
Pathology & Anatomical Regions \\
\midrule
Atelectasis & Left Lung, Right Lung \\
Cardiomegaly & Heart \\
Consolidation & Left Lung, Right Lung \\
Edema & Left Lung, Right Lung \\
Effusion & Left Lung, Right Lung \\
Emphysema & Left Lung, Right Lung \\
Fibrosis & Left Lung, Right Lung \\
Hernia & Facies Diaphragmatica, Mediastinum \\
Infiltration & Left Lung, Right Lung \\
Mass & Left Lung, Right Lung, Mediastinum \\
Nodule & Left Lung, Right Lung \\
Pleural Thickening & Left Lung, Right Lung \\
Pneumonia & Left Lung, Right Lung \\
Pneumothorax & Left Lung, Right Lung \\
\bottomrule
\end{tabular}
\end{table}
\appendixfloatbarrier
\section{Additional results}\label{appx:results-extra}
\suppressappendixfloats
\subsection{Localization and classification details}
At matched resolution, when macro IoU accuracy is compared at a T(IoU) in $\{0.1,\allowbreak0.2,\allowbreak0.3,\allowbreak0.4,\allowbreak 0.5\}$, the relative improvement over the published ThoraX-PriorNet-224 values is 17.1\%, 31.0\%, 32.9\%, 35.5\%, and 50.7\%, respectively. All relative changes against ThoraX-PriorNet in this section are computed from unrounded CLARiTy values and the published ThoraX-PriorNet values as printed, so recomputing them from the three-decimal CLARiTy values can differ by 0.1 percentage point. Localization stays high at the strictest threshold, $\text{T(IoU)} = 0.5$. The improvement likely stems from the multiple class tokens, which let the transformer allocate a separate representation to each class. This makes the attention maps class-specific.

\begin{table*}[!t]
\centering
\setlength{\abovecaptionskip}{0pt}
\caption{Classification performance comparison across different models evaluated on the NIH test set. The highest values are shown in bold for models that classify all 14 pathologies. The results of cited models are taken from their publications.}
\label{tab:classification_comparison}
\begin{threeparttable}
\renewcommand{\textsuperscript}[1]{\raisebox{0.45ex}{#1}}
\scriptsize
\setlength{\tabcolsep}{1.2pt}
\setlength{\aboverulesep}{0.5pt}
\setlength{\belowrulesep}{0.5pt}
\renewcommand{\arraystretch}{0.92}
\begin{tabular*}{\textwidth}{@{\extracolsep{\fill}} l *{15}{S[table-format=1.3,detect-weight=true,table-space-text-post={\textsuperscript{\textdagger}}]}@{}}
\toprule
{Model} & \multicolumn{15}{c}{Pathology AUC} \\
\cmidrule{2-16}
& {Atel.} & {Card.} & {Cons.} & {Edem.} & {Effu.} & {Emph.} & {Fibr.} & {Hern.} & {Infi.} & {Mass} & {Nodu.} & {Pleu.} & {Pn. 1} & {Pn. 2} & {Mean} \\
\midrule
\citet{Yao2017}\textsuperscript{\ddag} & 0.772 & 0.904 & 0.788 & 0.882 & 0.859 & 0.829 & 0.767 & 0.914 & 0.695 & 0.792 & 0.717 & 0.765 & 0.713 & 0.841 & 0.798 \\
\citet{Li2018}\textsuperscript{\ddag} & 0.800 & 0.870 & 0.800 & 0.880 & 0.870 & 0.910 & 0.780 & 0.770 & 0.700 & 0.830 & 0.750 & 0.790 & 0.670 & 0.870 & 0.806 \\
\citet{Ho2019}\textsuperscript{\ddag} & 0.795 & 0.887 & 0.786 & 0.892 & 0.875 & 0.875 & 0.756 & 0.836 & 0.703 & 0.835 & 0.716 & 0.774 & 0.742 & 0.863 & 0.810 \\
\citet{Ho2019}\textsuperscript{\P} & 0.726 & 0.865 & 0.717 & 0.812 & 0.810 & 0.740 & 0.748 & 0.803 & 0.682 & 0.773 & 0.686 & 0.730 & 0.660 & 0.796 & 0.753 \\
RGT\textsuperscript{1}\,\textsuperscript{\ddag} & 0.800\textsuperscript{\textdagger} & 0.920\textsuperscript{\textdagger} & \text{--} & \text{--} & 0.780\textsuperscript{\textdagger} & \text{--} & \text{--} & \text{--} & 0.860\textsuperscript{\textdagger} & 0.880\textsuperscript{\textdagger} & 0.880\textsuperscript{\textdagger} & \text{--} & 0.790\textsuperscript{\textdagger} & 0.810\textsuperscript{\textdagger} & 0.839\textsuperscript{\textdagger} \\
MLRFNet\textsuperscript{2}\,\textsuperscript{\ddag} & \bfseries 0.833 & 0.915 & \bfseries 0.826 & 0.905 & 0.884 & 0.941 & 0.821 & \bfseries 0.963 & 0.717 & 0.858 & 0.799 & 0.810 & 0.760 & 0.900 & \bfseries 0.853 \\
MLRFNet\textsuperscript{\P} & 0.756 & 0.888 & 0.742 & 0.840 & 0.821 & 0.894 & 0.816 & 0.910 & 0.684 & 0.808 & 0.735 & 0.760 & 0.711 & 0.850 & 0.801 \\
\citet{Kufel2023}\textsuperscript{\ddag} & 0.817 & 0.911 & 0.815 & 0.908 & 0.879 & 0.935 & 0.824 & 0.890 & 0.716 & 0.853 & 0.771 & 0.812 & 0.769 & 0.898 & 0.843 \\
\citet{Kufel2023}\textsuperscript{\P} & 0.766 & 0.863 & 0.736 & 0.846 & 0.830 & 0.926 & 0.818 & 0.846 & 0.699 & 0.794 & 0.740 & 0.776 & 0.725 & 0.873 & 0.803 \\
ThoraX-PriorNet-224\textsuperscript{3}\,\textsuperscript{\ddag} & 0.825 & 0.906 & 0.819 & 0.910 & 0.884 & 0.924 & 0.818 & 0.919 & 0.723 & 0.864 & 0.780 & 0.800 & 0.770 & 0.880 & 0.844 \\
ThoraX-PriorNet-512\textsuperscript{3}\,\textsuperscript{\ddag} & 0.827 & 0.902 & 0.812 & 0.908 & 0.884 & 0.927 & 0.826 & 0.905 & 0.723 & 0.867 & 0.807 & 0.813 & 0.764 & 0.890 & 0.847 \\
\citet{HungNguyen2024}\textsuperscript{\ddag} & 0.803 & 0.906 & 0.776 & 0.874 & 0.851 & 0.940 & 0.842 & 0.915 & 0.735 & 0.858 & 0.790 & 0.789 & 0.746 & 0.887 & 0.837 \\
\citet{HungNguyen2024}\textsuperscript{\P} & 0.768 & 0.895 & 0.744 & 0.839 & 0.827 & 0.902 & 0.814 & 0.898 & 0.700 & 0.816 & 0.753 & 0.768 & 0.709 & 0.855 & 0.806 \\
HydraViT\textsuperscript{4}\,\textsuperscript{\ddag} & 0.810 & 0.904 & 0.822 & 0.882 & 0.878 & 0.908 & 0.845 & 0.908 & \bfseries 0.755 & 0.840 & 0.800 & \bfseries 0.830 & 0.758 & 0.876 & 0.841 \\
HydraViT\textsuperscript{\P} & 0.732 & 0.876 & 0.720 & 0.822 & 0.804 & 0.889 & 0.795 & 0.844 & 0.695 & 0.776 & 0.690 & 0.736 & 0.679 & 0.837 & 0.778 \\
\citet{Li2026-CDCL}\textsuperscript{\ddag} & \bfseries 0.833 & 0.915 & 0.813 & \bfseries 0.918 & \bfseries 0.900 & 0.925 & 0.823 & 0.961 & 0.710 & 0.857 & 0.794 & 0.810 & 0.757 & 0.871 & 0.849 \\
CR-MSNet\textsuperscript{5}\,\textsuperscript{\ddag} & 0.827 & \bfseries 0.921 & 0.822 & 0.907 & 0.893 & 0.930 & 0.826 & 0.883 & 0.721 & \bfseries 0.876 & 0.802 & 0.801 & \bfseries 0.882 & 0.761 & 0.847 \\
\midrule
\citet{Wang2017-NIH} & 0.700 & 0.810 & 0.703 & 0.805 & 0.759 & 0.833 & 0.786 & 0.872 & 0.661 & 0.693 & 0.669 & 0.684 & 0.658 & 0.799 & 0.745 \\
\citet{YaoMar2018} & 0.733 & 0.856 & 0.711 & 0.806 & 0.806 & 0.842 & 0.743 & 0.775 & 0.673 & 0.777 & 0.718 & 0.724 & 0.684 & 0.805 & 0.761 \\
\citet{Tang2018} & 0.756 & 0.887 & 0.728 & 0.848 & 0.819 & 0.908 & 0.818 & 0.875 & 0.689 & 0.814 & 0.755 & 0.765 & 0.729 & 0.850 & 0.803 \\
\citet{Gundel2019} & 0.767 & 0.883 & 0.745 & 0.835 & 0.828 & 0.895 & 0.818 & 0.896 & 0.709 & 0.821 & 0.758 & 0.761 & 0.731 & 0.846 & 0.807 \\
\citet{Baltruschat2019} & 0.763 & 0.875 & 0.749 & 0.846 & 0.822 & 0.895 & 0.816 & 0.937 & 0.694 & 0.820 & 0.747 & 0.763 & 0.714 & 0.840 & 0.806 \\
SDFN\textsuperscript{6} & 0.781 & 0.885 & 0.743 & 0.842 & 0.832 & 0.921 & 0.835 & 0.911 & 0.700 & 0.815 & 0.765 & 0.791 & 0.719 & 0.866 & 0.815 \\
\citet{Ma2019} & 0.777 & 0.894 & 0.750 & 0.846 & 0.829 & 0.908 & 0.827 & 0.934 & 0.696 & 0.838 & 0.771 & 0.779 & 0.722 & 0.862 & 0.817 \\
Thorax-Net\textsuperscript{7} & 0.751 & 0.871 & 0.742 & 0.835 & 0.818 & 0.843 & 0.804 & 0.902 & 0.682 & 0.799 & 0.715 & 0.746 & 0.694 & 0.825 & 0.788 \\
CRAL\textsuperscript{8} & 0.781 & 0.880 & 0.754 & 0.850 & 0.829 & 0.908 & 0.830 & 0.917 & 0.702 & 0.834 & 0.773 & 0.778 & 0.729 & 0.857 & 0.816 \\
\citet{Chen2020} & 0.786 & 0.893 & 0.751 & 0.850 & 0.832 & \bfseries 0.944 & 0.834 & 0.929 & 0.699 & 0.840 & 0.800 & 0.795 & 0.739 & 0.876 & 0.826 \\
\citet{Guan2021} & 0.785 & 0.899 & 0.763 & 0.850 & 0.835 & 0.924 & 0.831 & 0.922 & 0.699 & 0.838 & 0.775 & 0.776 & 0.738 & 0.871 & 0.822 \\
\citet{Ouyang2021} & 0.770 & 0.870 & 0.740 & 0.840 & 0.830 & 0.940 & 0.830 & 0.910 & 0.710 & 0.830 & 0.790 & 0.790 & 0.720 & 0.880 & 0.819 \\
SwinCheX\textsuperscript{9} & 0.781 & 0.875 & 0.748 & 0.848 & 0.824 & 0.914 & 0.826 & 0.855 & 0.701 & 0.822 & 0.780 & 0.778 & 0.713 & 0.871 & 0.810 \\
\citet{Li-Zhou2022} & 0.797 & 0.872 & 0.779 & 0.858 & 0.852 & 0.935 & 0.825 & 0.907 & 0.711 & 0.843 & 0.795 & 0.800 & 0.742 & 0.894 & 0.829 \\
PCAN\textsuperscript{10} & 0.791 & 0.887 & 0.759 & 0.854 & 0.841 & \bfseries 0.944 & 0.819 & 0.928 & 0.711 & 0.839 & \bfseries 0.809 & 0.806 & 0.746 & 0.881 & 0.830 \\
TransDD\textsuperscript{11} & 0.791 & 0.885 & 0.753 & 0.859 & 0.842 & \bfseries 0.944 & 0.849 & 0.924 & 0.715 & 0.837 & 0.803 & 0.803 & 0.745 & 0.885 & 0.831 \\
Med-DGTN\textsuperscript{12} & 0.829 & 0.907 & 0.807 & 0.881 & 0.875 & 0.922 & 0.813 & 0.945 & 0.739 & 0.851 & 0.770 & 0.755 & 0.781 & 0.901 & 0.841 \\
\citet{Shi2026-GraphCrossAttn} & 0.803 & 0.913 & 0.790 & 0.902 & 0.843 & 0.936 & \bfseries 0.856 & 0.948 & 0.742 & 0.863 & 0.806 & 0.812 & 0.781 & \bfseries 0.903 & 0.849 \\
\midrule
ConvNeXtV2-B-224 & 0.772 & 0.892 & 0.750 & 0.853 & 0.828 & 0.900 & 0.809 & 0.894 & 0.699 & 0.814 & 0.766 & 0.770 & 0.715 & 0.859 & 0.809 \\
ConvNeXtV2-B-512 & 0.787 & 0.898 & 0.758 & 0.855 & 0.832 & 0.938 & 0.842 & 0.912 & 0.701 & 0.842 & 0.798 & 0.786 & 0.731 & 0.880 & 0.826 \\
CLARiTy-S-16-224 & 0.765 & 0.902 & 0.760 & 0.852 & 0.833 & 0.895 & 0.813 & 0.900 & 0.700 & 0.813 & 0.739 & 0.773 & 0.722 & 0.851 & 0.808 \\
CLARiTy-S-16-512 & 0.791 & 0.902 & 0.764 & 0.856 & 0.837 & 0.936 & 0.844 & 0.891 & 0.707 & 0.836 & 0.785 & 0.794 & 0.737 & 0.875 & 0.825 \\
\bottomrule
\end{tabular*}
\begin{tablenotes}
\footnotesize
\item Pathologies are: Atelectasis, Cardiomegaly, Consolidation, Edema, Effusion, Emphysema, Fibrosis, Hernia, Infiltration, Mass, Nodule, Pleural Thickening, Pneumonia, and Pneumothorax.
\item \textsuperscript{\P} Reproduced using the official NIH dataset split.
\item \textsuperscript{\textdagger} Classification performance evaluated on 8 pathologies: Atelectasis, Cardiomegaly, Effusion, Infiltration, Mass, Nodule, Pneumonia, and Pneumothorax.
\item \textsuperscript{\ddag} We could not verify that published model results used the official NIH dataset split.
\item ConvNeXtV2-B-224/512 are the CLARiTy teacher networks (\cref{sec:training}).
\item \textsuperscript{1} \citet{Han2023-RGT} \quad \textsuperscript{2} \citet{Li2023-MLRFNet} \quad \textsuperscript{3} \citet{Hossain2024-ThoraX-PriorNet} \quad \textsuperscript{4} \citet{Ozturk2025-HydraViT} \quad \textsuperscript{5} \citet{Wang2026-CRMSNet} \quad \textsuperscript{6} \citet{Liu2019} \quad \textsuperscript{7} \citet{Wang2019-Thorax-Net} \quad \textsuperscript{8} \citet{Guan2020} \quad \textsuperscript{9} \citet{Taslimi2022-SwinCheX} \quad \textsuperscript{10} \citet{Zhu2022-PCAN} \quad \textsuperscript{11} \citet{Jiang2024-TransDD} \quad \textsuperscript{12} \citet{Zhang2025-MedDGTN}
\end{tablenotes}
\end{threeparttable}
\end{table*}

\begin{table*}[!t]
\centering
\caption{Weakly-supervised localization performance comparison across different models on the bounding-box-annotated images of the NIH test set. For the CLARiTy models, localization is computed on all images of the localization test split, irrespective of whether the pathology was correctly classified. Localization results for the CLARiTy models are obtained with the calibrated OneBox extractor. The highest values for each T(IoU) are shown in bold. The results of cited models are taken from their publications.}
\label{tab:localization_comparison}
\setlength{\tabcolsep}{2pt}
\begin{threeparttable}
\footnotesize
\renewcommand{\arraystretch}{0.88}
\begin{tabular*}{\textwidth}{@{\extracolsep{\fill}} l l *{9}{S[table-format=1.3,detect-weight=true]}@{}}
\toprule
{T(IoU)} & {Model} & \multicolumn{9}{c}{Pathology IoU accuracy} \\
\cmidrule{3-11}
& & {Atel.} & {Card.} & {Effu.} & {Infi.} & {Mass} & {Nodu.} & {Pn. 1} & {Pn. 2} & {Mean} \\
\midrule
\multirow{9}{*}[-4pt]{0.1} & RGT\textsuperscript{1}\textsuperscript{\ddag} & 0.610 & 0.950 & 0.650 & 0.820 & 0.500 & 0.130 & 0.790 & 0.280 & 0.591 \\
 & ThoraX-PriorNet-224\textsuperscript{2}\textsuperscript{\ddag} & 0.628 & \bfseries 1.000 & 0.791 & 0.862 & 0.518 & 0.127 & 0.825 & 0.577 & 0.666 \\
 & ThoraX-PriorNet-512\textsuperscript{2}\textsuperscript{\ddag} & 0.761 & \bfseries 1.000 & 0.837 & 0.870 & 0.741 & 0.582 & 0.867 & \bfseries 0.763 & 0.803 \\
 & \citet{Wang2024-BAS}\textsuperscript{\ddag} & 0.250 & 0.990 & 0.490 & 0.630 & 0.270 & 0.030 & 0.680 & 0.300 & 0.450 \\
\addlinespace[0.2ex]
\cdashline{2-11}[1pt/1.5pt]
\addlinespace[0.4ex]
 & \citet{Wang2017-NIH} & 0.689 & 0.938 & 0.660 & 0.707 & 0.400 & 0.139 & 0.633 & 0.378 & 0.568 \\
 & \citet{Li-Zhou2022} & 0.635 & \bfseries 1.000 & 0.748 & 0.788 & 0.694 & 0.070 & 0.786 & 0.394 & 0.639 \\
 & PCAN\textsuperscript{3} & \bfseries 0.839 & \bfseries 1.000 & \bfseries 0.856 & \bfseries 0.935 & 0.824 & 0.494 & 0.900 & 0.375 & 0.778 \\
\addlinespace[0.2ex]
\cdashline{2-11}[1pt/1.5pt]
\addlinespace[0.4ex]
 & CLARiTy-S-16-224 & 0.780 & \bfseries 1.000 & 0.792 & 0.855 & 0.767 & 0.487 & \bfseries 0.966 & 0.592 & 0.780 \\
 & CLARiTy-S-16-512 & 0.769 & \bfseries 1.000 & 0.818 & 0.919 & \bfseries 0.837 & \bfseries 0.667 & 0.932 & 0.633 & \bfseries 0.822 \\
\midrule
\multirow{8}{*}[-4pt]{0.2} & RGT\textsuperscript{1}\textsuperscript{\ddag} & 0.410 & 0.910 & 0.410 & 0.590 & 0.260 & 0.050 & 0.570 & 0.190 & 0.424 \\
 & ThoraX-PriorNet-224\textsuperscript{2}\textsuperscript{\ddag} & 0.461 & \bfseries 1.000 & 0.621 & 0.626 & 0.318 & 0.013 & 0.692 & 0.340 & 0.509 \\
 & ThoraX-PriorNet-512\textsuperscript{2}\textsuperscript{\ddag} & 0.567 & 0.897 & \bfseries 0.693 & 0.724 & 0.577 & 0.253 & 0.692 & \bfseries 0.608 & 0.626 \\
\addlinespace[0.2ex]
\cdashline{2-11}[1pt/1.5pt]
\addlinespace[0.4ex]
 & \citet{Wang2017-NIH} & 0.472 & 0.685 & 0.451 & 0.478 & 0.259 & 0.051 & 0.350 & 0.235 & 0.373 \\
 & \citet{Li-Zhou2022} & 0.404 & \bfseries 1.000 & 0.664 & 0.737 & 0.429 & 0.014 & 0.691 & 0.277 & 0.527 \\
 & PCAN\textsuperscript{3} & \bfseries 0.650 & 0.760 & 0.575 & 0.789 & 0.612 & 0.190 & 0.767 & 0.250 & 0.574 \\
\addlinespace[0.2ex]
\cdashline{2-11}[1pt/1.5pt]
\addlinespace[0.4ex]
 & CLARiTy-S-16-224 & 0.626 & \bfseries 1.000 & 0.623 & 0.758 & 0.651 & 0.385 & \bfseries 0.881 & 0.408 & 0.667 \\
 & CLARiTy-S-16-512 & 0.593 & \bfseries 1.000 & 0.545 & \bfseries 0.823 & \bfseries 0.698 & \bfseries 0.667 & 0.847 & 0.408 & \bfseries 0.698 \\
\midrule
\multirow{9}{*}[-4pt]{0.3} & RGT\textsuperscript{1}\textsuperscript{\ddag} & 0.280 & 0.790 & 0.220 & 0.380 & 0.120 & 0.010 & 0.410 & 0.050 & 0.283 \\
 & ThoraX-PriorNet-224\textsuperscript{2}\textsuperscript{\ddag} & 0.300 & 0.986 & 0.458 & 0.439 & 0.212 & 0.000 & 0.542 & 0.247 & 0.398 \\
 & ThoraX-PriorNet-512\textsuperscript{2}\textsuperscript{\ddag} & \bfseries 0.467 & 0.795 & \bfseries 0.490 & 0.463 & 0.506 & 0.165 & 0.558 & \bfseries 0.464 & 0.488 \\
 & \citet{Wang2024-BAS}\textsuperscript{\ddag} & 0.070 & 0.900 & 0.170 & 0.240 & 0.130 & 0.020 & 0.230 & 0.110 & 0.230 \\
\addlinespace[0.2ex]
\cdashline{2-11}[1pt/1.5pt]
\addlinespace[0.4ex]
 & \citet{Wang2017-NIH} & 0.244 & 0.459 & 0.301 & 0.276 & 0.153 & 0.038 & 0.167 & 0.133 & 0.221 \\
 & \citet{Li-Zhou2022} & 0.205 & \bfseries 1.000 & 0.441 & 0.525 & 0.265 & 0.000 & 0.548 & 0.192 & 0.397 \\
 & PCAN\textsuperscript{3} & 0.428 & 0.336 & 0.333 & 0.569 & 0.482 & 0.038 & 0.600 & 0.125 & 0.364 \\
\addlinespace[0.2ex]
\cdashline{2-11}[1pt/1.5pt]
\addlinespace[0.4ex]
 & CLARiTy-S-16-224 & 0.418 & \bfseries 1.000 & 0.403 & 0.581 & \bfseries 0.558 & 0.308 & \bfseries 0.678 & 0.286 & 0.529 \\
 & CLARiTy-S-16-512 & 0.407 & \bfseries 1.000 & 0.325 & \bfseries 0.629 & \bfseries 0.558 & \bfseries 0.564 & 0.627 & 0.306 & \bfseries 0.552 \\
\multirow{8}{*}[-4pt]{0.4} & RGT\textsuperscript{1}\textsuperscript{\ddag} & 0.170 & 0.540 & 0.130 & 0.180 & 0.070 & 0.010 & 0.260 & 0.020 & 0.173 \\
 & ThoraX-PriorNet-224\textsuperscript{2}\textsuperscript{\ddag} & 0.172 & 0.945 & 0.261 & 0.309 & 0.165 & 0.000 & 0.392 & 0.124 & 0.296 \\
 & ThoraX-PriorNet-512\textsuperscript{2}\textsuperscript{\ddag} & \bfseries 0.322 & 0.616 & \bfseries 0.294 & 0.268 & 0.377 & 0.051 & 0.400 & \bfseries 0.340 & 0.334 \\
\addlinespace[0.2ex]
\cdashline{2-11}[1pt/1.5pt]
\addlinespace[0.4ex]
 & \citet{Wang2017-NIH} & 0.094 & 0.281 & 0.203 & 0.122 & 0.071 & 0.013 & 0.075 & 0.071 & 0.116 \\
 & \citet{Li-Zhou2022} & 0.103 & 0.979 & 0.273 & \bfseries 0.465 & 0.184 & 0.000 & 0.381 & 0.128 & 0.314 \\
 & PCAN\textsuperscript{3} & 0.228 & 0.096 & 0.163 & 0.366 & 0.365 & 0.013 & 0.325 & 0.104 & 0.207 \\
\addlinespace[0.2ex]
\cdashline{2-11}[1pt/1.5pt]
\addlinespace[0.4ex]
 & CLARiTy-S-16-224 & 0.209 & \bfseries 1.000 & 0.221 & 0.387 & 0.488 & 0.231 & 0.508 & 0.163 & 0.401 \\
 & CLARiTy-S-16-512 & 0.187 & \bfseries 1.000 & 0.221 & 0.339 & \bfseries 0.512 & \bfseries 0.385 & \bfseries 0.525 & 0.184 & \bfseries 0.419 \\
\midrule
\multirow{8}{*}[-4pt]{0.5} & RGT\textsuperscript{1}\textsuperscript{\ddag} & 0.080 & 0.320 & 0.050 & 0.090 & 0.050 & 0.000 & 0.120 & 0.010 & 0.090 \\
 & ThoraX-PriorNet-224\textsuperscript{2}\textsuperscript{\ddag} & 0.106 & 0.726 & 0.124 & 0.252 & 0.118 & 0.000 & 0.200 & 0.083 & 0.201 \\
 & ThoraX-PriorNet-512\textsuperscript{2}\textsuperscript{\ddag} & \bfseries 0.172 & 0.390 & \bfseries 0.144 & 0.179 & 0.294 & 0.013 & 0.317 & \bfseries 0.227 & 0.217 \\
\addlinespace[0.2ex]
\cdashline{2-11}[1pt/1.5pt]
\addlinespace[0.4ex]
 & \citet{Wang2017-NIH} & 0.050 & 0.178 & 0.111 & 0.065 & 0.012 & 0.013 & 0.033 & 0.031 & 0.062 \\
 & \citet{Li-Zhou2022} & 0.045 & 0.873 & 0.133 & \bfseries 0.343 & 0.123 & 0.000 & 0.333 & 0.096 & 0.243 \\
 & PCAN\textsuperscript{3} & 0.094 & 0.007 & 0.105 & 0.187 & 0.211 & 0.000 & 0.167 & 0.052 & 0.103 \\
\addlinespace[0.2ex]
\cdashline{2-11}[1pt/1.5pt]
\addlinespace[0.4ex]
 & CLARiTy-S-16-224 & 0.121 & \bfseries 1.000 & 0.078 & 0.274 & \bfseries 0.326 & 0.154 & \bfseries 0.390 & 0.082 & \bfseries 0.303 \\
 & CLARiTy-S-16-512 & 0.077 & \bfseries 1.000 & 0.117 & 0.226 & 0.302 & \bfseries 0.231 & 0.305 & 0.102 & 0.295 \\
\bottomrule
\end{tabular*}
\begin{tablenotes}
\item Pathologies are: Atelectasis, Cardiomegaly, Effusion, Infiltration, Mass, Nodule, Pneumonia, and Pneumothorax.
\item Median (mean) ground-truth box area (\% of image area): Nodu.~0.5 (0.6), Mass~1.6 (5.1), Atel.~2.0 (3.4), Pn.~2~3.1 (5.2), Effu.~5.0 (6.5), Pn.~1~6.3 (9.2), Infi.~6.4 (10.7), Card.~17.4 (17.9).
\item \textsuperscript{\ddag} We could not verify that published model results used the official NIH dataset split.
\item \textsuperscript{1} \citet{Han2023-RGT} \quad \textsuperscript{2} \citet{Hossain2024-ThoraX-PriorNet} \quad \textsuperscript{3} \citet{Zhu2022-PCAN}
\end{tablenotes}
\end{threeparttable}
\end{table*}

Compared with models operating at a higher resolution of $512 \times 512$, CLARiTy-S-16-224 remains competitive and even surpasses them in localization accuracy at higher T(IoU) thresholds. Its relative performance against ThoraX-PriorNet-512 is $-2.9$\%, 6.5\%, 8.4\%, 20.0\%, and 39.6\% as T(IoU) increases from 0.1 to 0.5. The comparison at $\text{T(IoU)} = 0.5$ includes the smallest annotated pathologies: nodules and masses have median ground-truth box areas of 0.5\% and 1.6\% of the image, compared with 2.0--17.4\% for the other six findings. At this threshold the 224 px model reaches IoU accuracies of 0.154 for nodules and 0.326 for masses; no method in the compared published results exceeds 0.013 on nodules, and the best compared 512 px model reaches 0.294 on masses. From $\text{T(IoU)} = 0.2$ onward the 224 px model exceeds every compared 512 px model on both findings. At the loosest threshold, $\text{T(IoU)} = 0.1$, the compared models lead on both. Against its own 512 px counterpart, CLARiTy-S-16-224 trails by 0.02--0.04 macro IoU accuracy at $\text{T(IoU)} = 0.1$--$0.4$ (\cref{tab:localization_comparison}).

The relative improvement in macro IoU accuracy of CLARiTy-S-16-512 over ThoraX-PriorNet-512 is 2.4\%, 11.4\%, 13.1\%, 25.4\%, and 35.9\% as T(IoU) increases from 0.1 to 0.5. The advantage therefore grows as the localization criterion becomes stricter.

To compare the two variants, we drew $2{,}000$ stratified bootstrap resamples. The macro IoU accuracy at $\text{T(IoU)} = 0.5$ is 0.303 (95\% confidence interval 0.271--0.333) for CLARiTy-S-16-224 and 0.295 (95\% confidence interval 0.263--0.331) for CLARiTy-S-16-512, so the two intervals overlap. \Cref{appx:bootstrap} lists the marginal intervals at every reported threshold.

Box counts, a matched-budget comparison with the original localization baseline of \citet{Wang2017-NIH}, and the stability of the OneBox calibration under refitting are reported in \cref{appx:onebox-robustness}.

CLARiTy-S-16-224 achieves classification results typical of prior work on the official NIH split, with a macro AUC of 0.808 (95\% confidence interval 0.804--0.812) within the 0.745--0.849 range reported for official-split models. It matches the classification performance of the ConvNeXtV2-B-224 teacher model (macro AUC of 0.808 versus 0.809). CLARiTy-S-16-512 achieves competitive classification results compared with prior work on the official NIH split, with a macro AUC of 0.825 (95\% confidence interval 0.821--0.829), also within the 0.745--0.849 range reported for official-split models. It matches the ConvNeXtV2-B-512 teacher (macro AUC of 0.825 versus 0.826). CLARiTy-S-16-512 improves on CLARiTy-S-16-224 by 1.7\% absolute macro AUC; the two variants also differ in initialization and in $p$, so this gain cannot be attributed to resolution alone.

\subsection{Computing resources}
The computing resources required for the CLARiTy-S-16-224 and CLARiTy-S-16-512 models and their components are given in \cref{tab:computing_resources}. For the CLARiTy-S-16-224 model, the combination of the SegmentCAM and attention pooling modules increases the number of parameters and FLOPS over the vision transformer backbone by 13.6\% and 17.6\%, respectively. For the CLARiTy-S-16-512 model, the increase in parameters and FLOPS is 13.4\% and 14.1\%, respectively.

\begin{table}[H]
\centering
\caption{Computing resources of the CLARiTy-S-16-224, CLARiTy-S-16-512, and their components.}
\label{tab:computing_resources}
\footnotesize
\setlength{\tabcolsep}{3pt}
\renewcommand{\arraystretch}{0.92}
\setlength{\belowcaptionskip}{2pt}
\begin{tabular}{@{}c l r r@{}}
\toprule
Image Size & Component & Parameters & FLOPS \\
\midrule
\multirow{4}{*}[-1.5pt]{224} & Vision Transformer Backbone & 21.68 M & 4.95 G \\
& Attention Pooling Module & 5.38 K & 5.38 K \\
& SegmentCAM Module & 2.94 M & 0.87 G \\
\cmidrule{2-4}
& CLARiTy-S-16-224 & 24.62 M & 5.83 G \\
\midrule
\multirow{4}{*}[-1.5pt]{512} & Vision Transformer Backbone & 21.99 M & 32.41 G \\
& Attention Pooling Module & 5.38 K & 5.38 K \\
& SegmentCAM Module & 2.94 M & 4.56 G \\
\cmidrule{2-4}
& CLARiTy-S-16-512 & 24.94 M & 36.98 G \\
\bottomrule
\end{tabular}
\end{table}

Example localization outputs of CLARiTy-S-16-224 are shown in \cref{fig:clarity-localization-outputs}, which spans \ifarxivbuild parts a to d\fi. The foreground masks confine the heatmaps and boxes to clinically relevant regions and improve accuracy.

\ifarxivbuild
\begin{figure*}[!t]
\centering
\begin{minipage}[t]{0.495\textwidth}
  \centering
  \includegraphics[width=\linewidth]{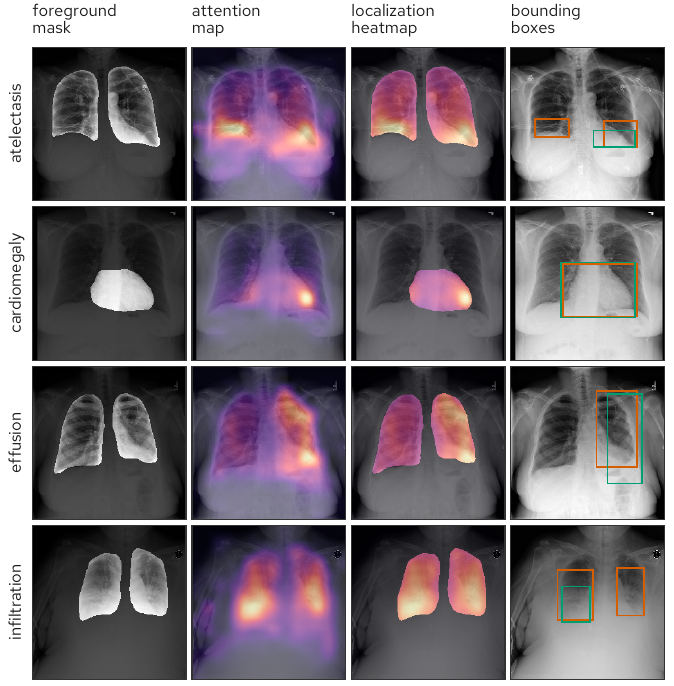}
  \par\smallskip
  {\footnotesize (a) Atelectasis, cardiomegaly, effusion and infiltration, first X-ray of each.}
\end{minipage}\hfill
\begin{minipage}[t]{0.495\textwidth}
  \centering
  \includegraphics[width=\linewidth]{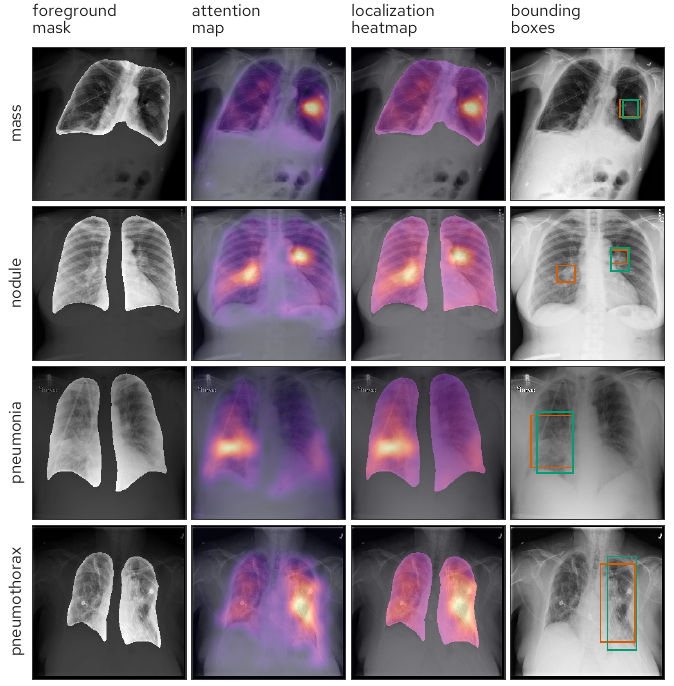}
  \par\smallskip
  {\footnotesize (b) Mass, nodule, pneumonia and pneumothorax, first X-ray of each.}
\end{minipage}

\smallskip
\begin{minipage}[t]{0.495\textwidth}
  \centering
  \includegraphics[width=\linewidth]{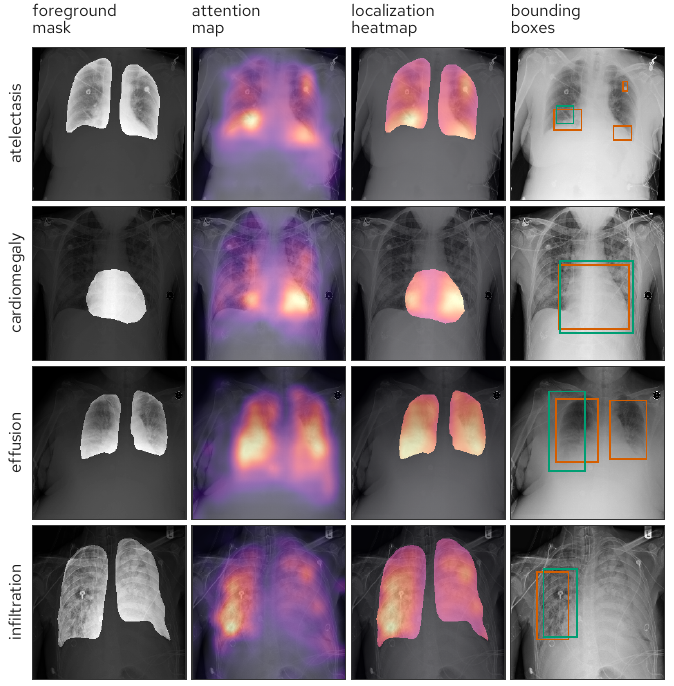}
  \par\smallskip
  {\footnotesize (c) Atelectasis, cardiomegaly, effusion and infiltration, second X-ray of each.}
\end{minipage}\hfill
\begin{minipage}[t]{0.495\textwidth}
  \centering
  \includegraphics[width=\linewidth]{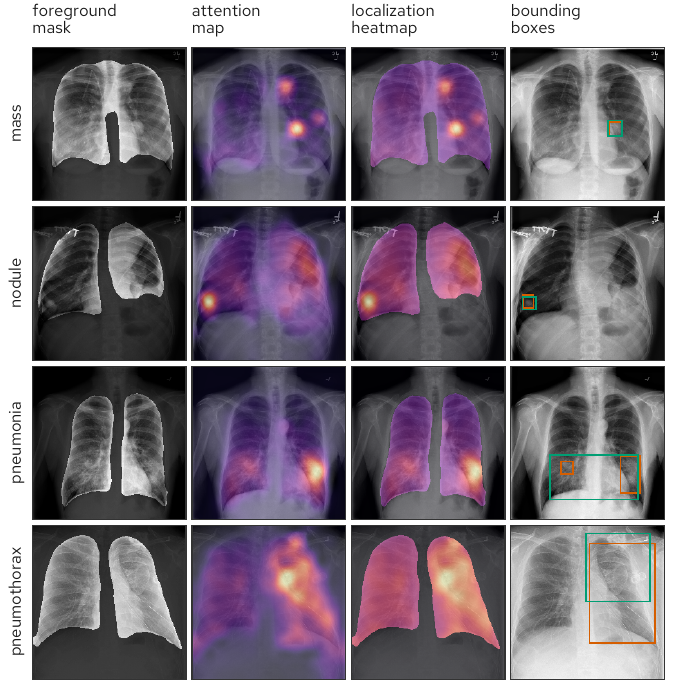}
  \par\smallskip
  {\footnotesize (d) Mass, nodule, pneumonia and pneumothorax, second X-ray of each.}
\end{minipage}
\caption{Selected localization outputs from CLARiTy-S-16-224 (parts a to d). For each pathology, two chest X-rays at $224 \times 224$ are shown: parts a and b use the first X-ray and parts c and d the second. All chest X-rays belong to the localization-test set. The model correctly predicted the presence of the pathology in each image. Class-specific outputs (columns): foreground mask, attention map, localization heatmap, and bounding boxes. The predicted bounding boxes are obtained with the calibrated OneBox extractor. Each output is overlaid on the input chest X-ray. Ground-truth bounding boxes are green and predicted bounding boxes vermilion.}
\label{fig:model-localization-outputs-a}
\label{fig:clarity-localization-outputs}
\label{fig:model-localization-outputs-b}
\label{fig:model-localization-outputs-c}
\label{fig:model-localization-outputs-d}
\end{figure*}
\fi

\FloatBarrier
\section{Ablation details}\label{appx:ablation}
\suppressappendixfloats
This section investigates the contributions of key components in the CLARiTy model through ablation studies on the SegmentCAM module, pretrained weights, classification loss, class token pooling, and the number of transformer layers whose attention maps are fused. Except for the attention-layer sweep, all ablations were performed on CLARiTy-S-16-224. For each variant, we selected the checkpoint with the highest validation UDP; the reported ablation results correspond to this checkpoint.

\subsection{SegmentCAM}
Two ablations were performed for the SegmentCAM module. Turning off the module completely (using only transformer self-attention for localization) caused relative drops in MSLI and UDP of 13.3\% and 3.9\%, respectively, while the validation macro AUC increased from 0.836 to 0.841 (\cref{tab:ablation_compact}). The module therefore contributes substantially to localization, at a small cost to classification performance. Because the foreground masks are derived from anatomical segmentations rather than pathology annotations, pathology-relevant regions outside the mapped anatomical regions can be suppressed, slightly reducing classification performance.

Removing the soft Dice segmentation loss caused relative drops in macro AUC, MSLI and UDP of 0.4\%, 22.8\% and 7.6\%, respectively, the largest localization drop of any ablation. The segmentation supervision contributes substantially to learning the correct foreground locations of each pathology.

\subsection{Pretrained weights}
An ablation was performed to investigate the impact of pretrained weights used for the CLARiTy-S-16-224. Replacing the DINO weights with ImageNet pretrained weights led to relative reductions in macro AUC (1.3\%), MSLI (7.0\%) and UDP (3.1\%), as shown in \cref{tab:ablation_compact}. The DINO weights thus give more discriminative patch embeddings, yielding more precise attention maps and higher classification accuracy.

\subsection{Classification loss}
An ablation experiment tested the commonly-used Multi-Label Soft Margin (MLSM) loss. It resulted in a slightly higher macro AUC (validation 0.841 versus 0.836) but relative drops in MSLI (2.3\%) and UDP (0.4\%) compared to the WBCE loss (\cref{tab:ablation_compact}). The UDP drop is the smallest among all ablations. The MLSM has the same definition as the WBCE in \labelcref{appx:wbcE}, except that $w_{\hspace{-0.08em} P} =\allowbreak w_{\hspace{-0.08em} N} =\allowbreak 1$. We therefore keep the dynamically weighted WBCE, which handles class imbalance in multi-label chest X-ray classification.

\subsection{Class token pooling}
An ablation was performed for the class token pooling method. The commonly-used GAP pooling yielded a slightly higher macro AUC than attention pooling (validation 0.843 versus 0.836), but the MSLI and UDP had relative drops of 6.4\% and 1.5\%, respectively (\cref{tab:ablation_compact}). The localization gain of attention pooling likely reflects the learned pooling weights, which let each class emphasize different embedding dimensions. With GAP pooling, all dimensions must have uniformly high activation values to positively classify a pathology \citep{Wang2017-NIH}. For classification, the choice of pooling has little impact.

\subsection{\texorpdfstring{Number of fused attention layers $p$}{Number of fused attention layers p}}
An ablation was performed on the number of transformer layers $p$ whose class-specific attention maps are fused, for both the CLARiTy-S-16-224 and the CLARiTy-S-16-512 variants. A sweep over all $p \in \{ 1,\allowbreak 2,\allowbreak \dots ,\allowbreak 12 \}$ was evaluated using the validation UDP, with each $p$ represented by its best validation checkpoint reached during training.

As shown in \cref{fig:p-max-val-UDP}, both variants follow a similar trend: the validation UDP initially increases as more attention layers are fused, reaches a maximum at a small number of fused layers, and then slowly declines as progressively earlier layers are included. For CLARiTy-S-16-224, the validation UDP increases from 0.594 at $p=1$ to a maximum of 0.614 at $p=5$, before decreasing to 0.608 at $p=12$. For CLARiTy-S-16-512, the validation UDP increases from 0.618 at $p=1$ to a maximum of 0.636 at $p=3$, remains close to this value up to $p=5$, and decreases to 0.626 at $p=12$. These results show that fusing the attention maps of a subset of the final transformer layers produces the most class-discriminative localization heatmaps, whereas including the earliest layers dilutes the fused attention map.
\appendixfloatbarrier
\section{Localization failure analysis}\label{appx:failures}
\suppressappendixfloats
This appendix characterizes the localization failures at $\text{T(IoU)} = 0.5$ for the CLARiTy-S-16-224 and CLARiTy-S-16-512 variants. Predicted boxes are those obtained by applying the calibrated OneBox extractor (\labelcref{appx:onebox-params}) to the fused heatmaps. \Cref{tab:failures_by_area} shows that detection accuracy increases with the ground-truth box area, from $0.06/0.17$ (224/512) below $0.5\%$ of the image to $0.64$ for both variants at $10\%$ or more. Of the 493 localization-test instances over 438 images, 154 (224) and 148 (512) are detected; \cref{tab:failures_per_finding} breaks the remainder down per finding. Most small lesions are missed: 70 (224) and 64 (512) of the 74 instances with boxes covering less than 1\% of the image area go unmatched. These account for about a fifth of all misses. Across all misses, 10 (224) and 4 (512) match the left--right mirror image of the ground-truth box. Unmatched extra boxes are common (mean 1.28 and 1.42 per instance for the 224 and 512 variants, present in 77\% and 80\% of instances), whereas boxes matching a different finding's ground-truth box in the 47 multi-finding images are rare (11 and 9 boxes in total). The two variants differ mainly in the two smallest area bands: the 512 px model detects more instances below $1\%$ of the image area ($0.167$ against $0.056$ in the smallest band and $0.105$ against $0.053$ in the next). The 224 px model is ahead between $2\%$ and $10\%$, and the two are nearly level in the largest band ($0.638$ against $0.644$).

\begin{table}[!htbp]
\centering
\caption{Detection accuracy at $\text{T(IoU)} = 0.5$ by ground-truth box area (fraction of the image), pooled over the 493 localization-test instances.}
\label{tab:failures_by_area}
\begin{tabular}{l r c c}
\toprule
{GT box area} & {Instances} & {224} & {512} \\
\midrule
$< 0.5\%$       & 36  & 0.056 & 0.167 \\
$0.5$--$1\%$    & 38  & 0.053 & 0.105 \\
$1$--$2\%$      & 75  & 0.173 & 0.173 \\
$2$--$5\%$      & 108 & 0.157 & 0.111 \\
$5$--$10\%$     & 87  & 0.287 & 0.195 \\
$\geq 10\%$     & 149 & 0.638 & 0.644 \\
\bottomrule
\end{tabular}
\end{table}

\begin{table*}[!t]
\centering
\caption{Localization failures at $\text{T(IoU)} = 0.5$ on all 493 localization-test instances, per variant. ``Missed'' counts instances without a matching prediction and is followed in parentheses by the number of misses with a small ground-truth box (less than 1\% of the image area) and with a mirrored prediction (matching the left--right mirrored ground-truth box). ``Extra boxes'' is the mean number of predicted boxes per instance that do not match the instance's ground-truth box.}
\label{tab:failures_per_finding}
\footnotesize
\begin{minipage}[t]{0.495\textwidth}
\centering
CLARiTy-S-16-224\\[3pt]
\begin{tabular*}{\linewidth}{@{\extracolsep{\fill}} l r c c r@{}}
\toprule
{Finding} & {Instances} & {Accuracy} & \makecell{Missed\\(small/mirr.)} & \makecell{Extra\\boxes} \\
\midrule
Atelectasis    & 91 & 0.121 & 80 (12/2) & 1.58 \\
Cardiomegaly   & 73 & 1.000 & 0 (0/0)   & 0.00 \\
Effusion       & 77 & 0.078 & 71 (7/4)  & 1.77 \\
Infiltration   & 62 & 0.274 & 45 (2/2)  & 1.35 \\
Mass           & 43 & 0.326 & 29 (10/0) & 1.09 \\
Nodule         & 39 & 0.154 & 33 (31/0) & 1.79 \\
Pneumonia      & 59 & 0.390 & 36 (0/1)  & 1.39 \\
Pneumothorax   & 49 & 0.082 & 45 (8/1)  & 1.43 \\
\bottomrule
\end{tabular*}
\end{minipage}\hfill
\begin{minipage}[t]{0.495\textwidth}
\centering
CLARiTy-S-16-512\\[3pt]
\begin{tabular*}{\linewidth}{@{\extracolsep{\fill}} l r c c r@{}}
\toprule
{Finding} & {Instances} & {Accuracy} & \makecell{Missed\\(small/mirr.)} & \makecell{Extra\\boxes} \\
\midrule
Atelectasis    & 91 & 0.077 & 84 (12/0) & 1.36 \\
Cardiomegaly   & 73 & 1.000 & 0 (0/0)   & 0.00 \\
Effusion       & 77 & 0.117 & 68 (7/2)  & 1.70 \\
Infiltration   & 62 & 0.226 & 48 (2/1)  & 1.55 \\
Mass           & 43 & 0.302 & 30 (9/0)  & 1.84 \\
Nodule         & 39 & 0.231 & 30 (26/0) & 2.44 \\
Pneumonia      & 59 & 0.305 & 41 (0/0)  & 1.44 \\
Pneumothorax   & 49 & 0.102 & 44 (8/1)  & 1.84 \\
\bottomrule
\end{tabular*}
\end{minipage}
\end{table*}

\appendixfloatbarrier
\section{Subgroup analysis}\label{appx:subgroups}
\suppressappendixfloats
\Cref{tab:subgroups} reports the classification macro AUC of the two variants by patient sex, patient age band, and radiograph view position on the official NIH test set ($25{,}595$ images). The results are descriptive: subgroup gaps are reported as observed, and the analysis is not powered for subgroup inference. The sex gap is negligible ($0.001$--$0.003$), the AP view is about $0.02$ below the posteroanterior (PA) view, and the largest spread across age bands ($0.05$--$0.06$) is driven by patients younger than 40 years ($n = 8{,}411$), for whom several pathologies are less frequent. The 75+ band is small ($n = 550$). Sex is the sex recorded in the NIH metadata.

\begin{table}[H]
\centering
\caption{Classification macro AUC by subgroup on the official NIH test set.}
\label{tab:subgroups}
\begin{tabular}{l r c c}
\toprule
{Group} & {Images} & {224} & {512} \\
\midrule
All            & 25{,}595 & 0.808 & 0.825 \\
\midrule
Male           & 14{,}881 & 0.809 & 0.825 \\
Female         & 10{,}714 & 0.806 & 0.824 \\
\midrule
PA view        & 11{,}096 & 0.803 & 0.822 \\
AP view        & 14{,}499 & 0.786 & 0.803 \\
\midrule
Age $< 40$     & 8{,}411  & 0.749 & 0.770 \\
Age $40$--$59$ & 11{,}222 & 0.805 & 0.821 \\
Age $60$--$74$ & 5{,}412  & 0.795 & 0.812 \\
Age $\geq 75$  & 550      & 0.798 & 0.814 \\
\bottomrule
\end{tabular}
\end{table}

\appendixfloatbarrier
\section{OneBox calibration details}\label{appx:onebox}
\suppressappendixfloats

\subsection{Search space}\label{appx:onebox-space}
OneBox has five parameters: the Gaussian blur width $\sigma$, the two thresholds $t_{\mathrm{lo}} < t_{\mathrm{hi}}$, the interpolation fraction $s$ between the core and support boxes (\cref{sec:onebox}), and the isotropic scale $\alpha$. Fitting evaluates a fixed grid per finding and image resolution: $\sigma$ over five values on a grid defined relative to the image side length ($\{0, 1, 2, 3, 4\}$ pixels at $224 \times 224$, scaled proportionally at other resolutions); $t_{\mathrm{lo}}$ and $t_{\mathrm{hi}}$ over all 171 ordered pairs on a $0.05$ grid between $0.05$ and $0.95$; $s$ over $\{0, 0.25, 0.5, 0.75, 1, 1.25\}$; and $\alpha$ over $\{0.9, 1.0, 1.1\}$. The gate keeps a support region only if it contains a core of at least four pixels, so the combinations amount to $15{,}390$ configurations per finding.

\subsection{Fitted parameters}\label{appx:onebox-params}
\Cref{tab:onebox_params} lists the parameters selected on the localization validation split for the two variants reported in this work. For each finding, the configuration that maximizes the MSLI over the search grid, with ties broken in favor of fewer boxes. The parameters are fitted independently for each image resolution and frozen before the test split is evaluated; the 224 and 512 values therefore need not agree.

\begin{table*}[!tb]
\centering
\caption{Calibrated OneBox parameters for the CLARiTy-S-16-224 and CLARiTy-S-16-512 variants. $\sigma$ is given in pixels at the model's input resolution; the search grid is relative to the image side length. The minimum gate area is four pixels for both variants.}
\label{tab:onebox_params}
\footnotesize
\begin{minipage}[t]{0.495\textwidth}
\centering
CLARiTy-S-16-224\\[3pt]
\begin{tabular*}{\linewidth}{@{\extracolsep{\fill}} l r r r r r@{}}
\toprule
{Finding} & {$\sigma$} & {$t_{\mathrm{lo}}$} & {$t_{\mathrm{hi}}$} & {$s$} & {$\alpha$} \\
\midrule
Atelectasis  & 0.00 & 0.75 & 0.80 & 0.25 & 1.10 \\
Cardiomegaly & 4.00 & 0.05 & 0.40 & 1.25 & 1.00 \\
Effusion     & 0.00 & 0.70 & 0.75 & 1.00 & 1.10 \\
Infiltration & 2.00 & 0.65 & 0.80 & 0.50 & 1.10 \\
Mass         & 1.00 & 0.70 & 0.75 & 0.50 & 1.10 \\
Nodule       & 2.00 & 0.75 & 0.80 & 0.75 & 0.90 \\
Pneumonia    & 3.00 & 0.50 & 0.55 & 0.50 & 0.90 \\
Pneumothorax & 3.00 & 0.45 & 0.70 & 1.25 & 0.90 \\
\bottomrule
\end{tabular*}
\end{minipage}\hfill
\begin{minipage}[t]{0.495\textwidth}
\centering
CLARiTy-S-16-512\\[3pt]
\begin{tabular*}{\linewidth}{@{\extracolsep{\fill}} l r r r r r@{}}
\toprule
{Finding} & {$\sigma$} & {$t_{\mathrm{lo}}$} & {$t_{\mathrm{hi}}$} & {$s$} & {$\alpha$} \\
\midrule
Atelectasis  & 0.00 & 0.35 & 0.80 & 0.25 & 1.10 \\
Cardiomegaly & 9.14 & 0.05 & 0.70 & 1.25 & 1.00 \\
Effusion     & 0.00 & 0.50 & 0.80 & 0.25 & 1.10 \\
Infiltration & 4.57 & 0.05 & 0.75 & 0.25 & 1.10 \\
Mass         & 0.00 & 0.55 & 0.65 & 1.25 & 1.10 \\
Nodule       & 0.00 & 0.65 & 0.70 & 1.25 & 1.10 \\
Pneumonia    & 0.00 & 0.45 & 0.75 & 0.25 & 1.10 \\
Pneumothorax & 2.29 & 0.50 & 0.60 & 0.50 & 1.00 \\
\bottomrule
\end{tabular*}
\end{minipage}
\end{table*}

Read across the two panels, many fits need no smoothing at all ($\sigma = 0$: two findings at 224 px and five at 512 px), and the isotropic scale is $\alpha = 1.10$ for four findings at 224 px and six at 512 px. The interpolation fraction $s = 1.25$ appears only for Cardiomegaly and Pneumothorax at 224 px and for Cardiomegaly, Mass and Nodule at 512 px. Cardiomegaly has the largest smoothing width in both fits, from $\sigma = 4.00$ px at 224 px to $9.14$ px at 512 px, the same fraction of the image side length at the two resolutions.
\appendixfloatbarrier
\section{Extractor budget and calibration stability}\label{appx:onebox-robustness}
\suppressappendixfloats

The OneBox fitting objective counts a detected instance as a success and applies no
false-positive penalty, and the selector does not cap the number of output
boxes. The reported localization accuracies could therefore be
inflated by returning more boxes than a fixed-threshold extractor.
\Cref{tab:appendix-extractor-budget} tests this possibility by reporting the
mean number of predicted boxes and the unmatched boxes at
$\text{T(IoU)} = 0.5$ together with the accuracy, and by comparing
against the localization baseline of \citet{Wang2017-NIH}. This baseline
thresholds each class heatmap per image at $0.235$ and $0.706$ ($60$ and
$180$ on a $0$--$255$ scale) and forms connected components, the rule
defined in the original dataset paper. It is applied to the same maps and instances and
capped to the same number of boxes per instance as OneBox (the box budget). Uncapped, the Wang
baseline returns 3.9 (224)
and 6.0 (512) boxes per
instance, with
3.5 and
5.6 unmatched at
$\text{T(IoU)} = 0.5$, versus
1.6 and
1.7 boxes with
1.3 and
1.4 unmatched for
OneBox. The uncapped Wang baseline leads at the coarser thresholds
($\text{T(IoU)} \leq 0.4$ for the 224 variant and $\leq 0.3$ for the
512 variant). At $\text{T(IoU)} = 0.5$ it is level with OneBox at 224 px (0.302 against 0.303) and behind it at 512 px (0.285 against 0.295), and it
emits several times as many unmatched boxes. At a matched budget,
OneBox ranks above the Wang baseline at every threshold and for both variants: with one box per instance it improves the
macro IoU accuracy over the Wang baseline by 0.12--0.19
(224) and 0.12--0.20 (512) across
$\text{T(IoU)} = 0.1$--$0.5$, and with two boxes by
0.04--0.10 and 0.04--0.08.
The low-threshold advantage of the uncapped baseline is therefore explained
by its larger box output, and the OneBox
gains are not an artifact of box inflation.

The calibration is fitted on a small validation sample
(40--89
bounding-box instances per finding; \labelcref{appx:onebox-space}), so we
verified that the reported accuracy does not depend on this particular
sample. The same 15{,}390 configuration grid was refitted 500 times on
bootstrap resamples of the per-finding validation instances and 200 times
on random 50/50 splits of them, fitting each half separately, and the test
set was scored with each resulting parameter set. In most refits the
selected parameters differ from those in \cref{tab:onebox_params}, yet
the resulting accuracy changes little. Refitting changes the mean test
macro IoU accuracy by at most 0.016 under the bootstrap draws and 0.020
under the split-half draws. The 2.5--97.5 percentile range across the five
IoU thresholds is at most 0.061 and 0.057, respectively.
The mean refitted accuracy lies inside the test-set bootstrap confidence
interval at the reported $\text{T(IoU)} = 0.1$ and $0.5$
(\cref{tab:appendix-localization-ci}). The reported localization results therefore do not depend on a particular fit,
and calibrating on the available validation split is sufficient for the
conclusions drawn here. The reported models use that full-validation fit
(\labelcref{appx:onebox-params}).

\begin{table*}[!tb]
\centering
\caption[Extractor comparison at matched box budgets]{Extractor comparison on the localization test split (438 images, 493 bounding-box instances), reporting macro IoU accuracy over the eight bounding-box findings, mean predicted boxes per instance, and unmatched boxes per instance at $\text{T(IoU)} = 0.5$ (a false-positive count). The localization baseline of \citet{Wang2017-NIH} (per-image thresholding at $0.235$ and $0.706$ with connected components) is evaluated on the same maps. Top-$k$ keeps the $k$ highest-peak boxes per instance. The matched-budget comparison sets OneBox against the Wang top-$k$ rows.}
\label{tab:appendix-extractor-budget}
\setlength{\tabcolsep}{0.6pt}
\begin{tabular*}{\textwidth}{@{\extracolsep{\fill}}l r r c c c c c}
\toprule
& & & \multicolumn{5}{c}{\makecell{macro IoU accuracy\\at T(IoU)}} \\
\cmidrule(lr){4-8}
Extractor & Boxes/inst & Unmatched@0.5 & 0.1 & 0.2 & 0.3 & 0.4 & 0.5 \\
\midrule
\multicolumn{8}{c}{CLARiTy-S-16-224} \\
\midrule
Wang & 3.89 & 3.55 & 0.830 & 0.730 & 0.590 & 0.424 & 0.302 \\
OneBox & 1.60 & 1.29 & 0.780 & 0.667 & 0.529 & 0.401 & 0.303 \\
Wang, top-2 & 2.00 & 1.71 & 0.672 & 0.582 & 0.467 & 0.353 & 0.255 \\
OneBox, top-2 & 1.52 & 1.21 & 0.768 & 0.658 & 0.521 & 0.397 & 0.298 \\
Wang, top-1 & 1.00 & 0.84 & 0.507 & 0.375 & 0.252 & 0.200 & 0.144 \\
OneBox, top-1 & 1.00 & 0.73 & 0.640 & 0.551 & 0.439 & 0.342 & 0.261 \\
\midrule
\multicolumn{8}{c}{CLARiTy-S-16-512} \\
\midrule
Wang & 5.99 & 5.61 & 0.881 & 0.727 & 0.570 & 0.413 & 0.285 \\
OneBox & 1.71 & 1.41 & 0.822 & 0.698 & 0.552 & 0.419 & 0.295 \\
Wang, top-2 & 2.00 & 1.67 & 0.737 & 0.622 & 0.485 & 0.361 & 0.258 \\
OneBox, top-2 & 1.58 & 1.28 & 0.813 & 0.691 & 0.556 & 0.422 & 0.298 \\
Wang, top-1 & 1.00 & 0.82 & 0.563 & 0.416 & 0.297 & 0.227 & 0.168 \\
OneBox, top-1 & 1.00 & 0.71 & 0.710 & 0.606 & 0.493 & 0.375 & 0.285 \\
\bottomrule
\end{tabular*}
\end{table*}
\appendixfloatbarrier
\section{Statistical uncertainty}\label{appx:bootstrap}
\suppressappendixfloats
\setlength{\floatsep}{10pt plus 2pt minus 2pt}
\setlength{\textfloatsep}{12pt plus 2pt minus 3pt}
Statistical uncertainty on the test-set metrics of the reported models is quantified with $2{,}000$ stratified image-level bootstrap resamples, which are used to form the localization intervals in \cref{tab:appendix-localization-ci} and the per-pathology AUC intervals in \cref{tab:appendix-auc-ci}. The intervals characterize the sampling uncertainty of the reported values; comparisons against the cited models rely on their published single values, for which per-image predictions are not available.

Each interval is the range that contains the middle 95\% of the bootstrap estimates, so it describes how much the reported value would move if the test images were resampled, not the probability that the true value lies above a fixed target. Because the resampling is at the image level and stratified, the intervals reflect variation between images rather than the number of boxes drawn on them. At $\text{T(IoU)} = 0.5$ the localization interval for the 224 px model spans 0.271 to 0.333, whereas the macro-AUC interval spans 0.804 to 0.812. The narrower AUC interval is expected from the much larger classification test split (25{,}595 images against 438). Overlapping intervals are not a test of a difference, so the comparisons with the published models rest on their single reported values.

\begin{table}[!tb]
\centering
\caption[Localization IoU accuracy: 95\% bootstrap confidence intervals]{95\% stratified percentile bootstrap confidence intervals for the macro IoU accuracy over the 8 bounding-box pathologies at each $\mathrm{T(IoU)}$ on the localization test split (438 images, 493 instances; 2{,}000 image-level resamples). Each cell gives the point estimate followed by the 95\% interval.}
\label{tab:appendix-localization-ci}
\setlength{\tabcolsep}{3pt}
\begin{tabular}{@{}lcc@{}}
\toprule
$\mathrm{T(IoU)}$ & CLARiTy-S-16-224 & CLARiTy-S-16-512 \\
\midrule
0.1 & 0.780 [0.745, 0.814] & 0.822 [0.789, 0.854] \\
0.2 & 0.667 [0.631, 0.705] & 0.698 [0.662, 0.733] \\
0.3 & 0.529 [0.491, 0.568] & 0.552 [0.516, 0.596] \\
0.4 & 0.401 [0.365, 0.436] & 0.419 [0.384, 0.459] \\
0.5 & 0.303 [0.271, 0.333] & 0.295 [0.263, 0.331] \\
\bottomrule
\end{tabular}
\end{table}
\begin{table*}[!t]
\centering
\caption[Per-pathology AUC: 95\% bootstrap confidence intervals]{95\% stratified percentile bootstrap confidence intervals for the per-pathology and macro AUC of CLARiTy-S-16-224, CLARiTy-S-16-512 and their ConvNeXtV2-B teachers on the official NIH test split (25{,}595 images; 2{,}000 image-level resamples). Each cell gives the point estimate followed by the 95\% interval.}
\label{tab:appendix-auc-ci}
\footnotesize
\setlength{\tabcolsep}{4pt}
\begin{tabular*}{\textwidth}{@{\extracolsep{\fill}} l c c c c@{}}
\toprule
Pathology & CLARiTy-S-16-224 & ConvNeXtV2-B-224 & CLARiTy-S-16-512 & ConvNeXtV2-B-512 \\
\midrule
Atelectasis & 0.765 [0.757, 0.773] & 0.772 [0.764, 0.780] & 0.791 [0.783, 0.798] & 0.787 [0.779, 0.794] \\
Cardiomegaly & 0.902 [0.894, 0.909] & 0.892 [0.884, 0.900] & 0.902 [0.894, 0.909] & 0.898 [0.890, 0.905] \\
Consolidation & 0.760 [0.751, 0.770] & 0.750 [0.740, 0.759] & 0.764 [0.754, 0.773] & 0.758 [0.749, 0.767] \\
Edema & 0.852 [0.843, 0.861] & 0.853 [0.844, 0.861] & 0.856 [0.847, 0.864] & 0.855 [0.846, 0.863] \\
Effusion & 0.833 [0.827, 0.839] & 0.828 [0.822, 0.834] & 0.837 [0.831, 0.843] & 0.832 [0.826, 0.837] \\
Emphysema & 0.895 [0.885, 0.904] & 0.900 [0.891, 0.909] & 0.936 [0.929, 0.942] & 0.938 [0.931, 0.944] \\
Fibrosis & 0.813 [0.796, 0.830] & 0.809 [0.791, 0.825] & 0.844 [0.829, 0.860] & 0.842 [0.827, 0.857] \\
Hernia & 0.900 [0.859, 0.935] & 0.894 [0.851, 0.931] & 0.891 [0.848, 0.931] & 0.912 [0.871, 0.946] \\
Infiltration & 0.700 [0.692, 0.706] & 0.699 [0.692, 0.706] & 0.707 [0.700, 0.714] & 0.701 [0.695, 0.708] \\
Mass & 0.813 [0.803, 0.823] & 0.814 [0.804, 0.823] & 0.836 [0.827, 0.845] & 0.842 [0.833, 0.851] \\
Nodule & 0.739 [0.728, 0.750] & 0.766 [0.755, 0.777] & 0.785 [0.774, 0.796] & 0.798 [0.788, 0.808] \\
Pleural Thickening & 0.773 [0.761, 0.785] & 0.770 [0.757, 0.781] & 0.794 [0.782, 0.805] & 0.786 [0.774, 0.798] \\
Pneumonia & 0.722 [0.705, 0.737] & 0.715 [0.698, 0.732] & 0.737 [0.722, 0.753] & 0.731 [0.716, 0.746] \\
Pneumothorax & 0.851 [0.844, 0.858] & 0.859 [0.852, 0.865] & 0.875 [0.869, 0.881] & 0.880 [0.874, 0.885] \\
\midrule
macro AUC & 0.808 [0.804, 0.812] & 0.809 [0.804, 0.812] & 0.825 [0.821, 0.829] & 0.826 [0.822, 0.829] \\
\bottomrule
\end{tabular*}
\end{table*}
\appendixfloatbarrier
\section{Cross-dataset transfer: additional material}\label{appx:transfer}
\suppressappendixfloats

The six model settings are the NIH-calibrated and target-calibrated CLARiTy models at 224 and 512 px and the two external models. The NIH-calibrated model applies the NIH checkpoint with its OneBox parameters fitted on NIH and frozen, and uses no target data. The target-calibrated model keeps the same weights and heatmaps and refits the five OneBox parameters per finding on target validation boxes, with no weight change. The two external models are RAD-DINO 518~\cite{PerezGarcia2025-RADDINO}, whose per-finding logistic probe is fitted on target training labels, and TorchXRayVision DenseNet121-NIH 224 (TXV-NIH)~\cite{Cohen2022-TorchXrayVision}, which fits no classifier. Each is read out with HiResCAM~\cite{Draelos2020-HiResCAM}, a gradient-based map, through its own OneBox extractor fitted on the target validation images.

\subsection{RSNA split: definition and provenance}
RSNA Pneumonia provides a different set of box annotations for images drawn from ChestX-ray14~\cite{Shih2019-RSNA,Wang2017-NIH}. The released RSNA test images carry no labels, so a stratified 70/10/20 partition of the labeled images would be the natural test set. Of the 5,338 test images that partition produces, however, 3,875 (72.6\%) and 688 of 1,203 positives (57.2\%) have their NIH source in the NIH training or validation split used to train the checkpoints. We therefore use one RSNA test set, the NIH-test-derived split of 7,186 images (2,589 positives, 4,229 boxes) whose NIH source is in the official ChestX-ray14 test list, which is 27\% of the 26,684 RSNA images. The remaining images are divided into training (87.5\%) and validation (12.5\%; 2,438 images, 428 positives) sets by stratified sampling. The validation set that fits the OneBox parameters is drawn from these remaining images; only those parameters are fitted, so the weights and heatmaps remain unchanged. Prevalence is 36.0\%, and the anteroposterior view share is 63.2\%.

\subsection{VinDr-CXR label map}
VinDr-CXR~\cite{Nguyen2022-VinDrCXR} has 22 box classes. Six map to NIH findings that carry NIH boxes and are scored for localization and classification: Atelectasis, Cardiomegaly, Pleural effusion, Infiltration, Nodule/Mass and Pneumothorax. Nodule/Mass is the only many-to-one case (VinDr has one class; NIH has Mass and Nodule), so the finding is the maximum of the two NIH outputs, taken on post-activation values at the model-output boundary. Five further classes (Consolidation, Edema, Emphysema, Pulmonary fibrosis and Pleural thickening) have NIH counterparts without NIH boxes and enter the classification macro only. Edema has no positive test image, so that macro covers 10 findings. The transfer claim therefore covers 11 of the 22 box classes: six for localization and five for classification. The other 11 box classes have no NIH counterpart (Aortic enlargement, Calcification, Clavicle fracture, Enlarged PA, ILD, Lung Opacity, Lung cavity, Lung cyst, Mediastinal shift, Rib fracture and Other lesion). They are reported with counts and never scored. Pneumonia is an image-level label with no VinDr boxes. It is never remapped (Lung Opacity is not Pneumonia) and never enters a VinDr macro. The only change to the region priors concerns pleural effusion. VinDr effusion boxes sit at the costophrenic angle, so Facies Diaphragmatica is added for VinDr only and the NIH region map is unchanged. At 512 px this raises the fraction of effusion boxes inside the mapped region on the test set from 0.327 to 0.754, and the six-finding macro share of boxes inside a mapped region from 0.775 to 0.838. All other region priors are unchanged.

\subsection{Per-finding and per-threshold results}
\begin{figure}[p]
\centering
\includegraphics[width=\linewidth]{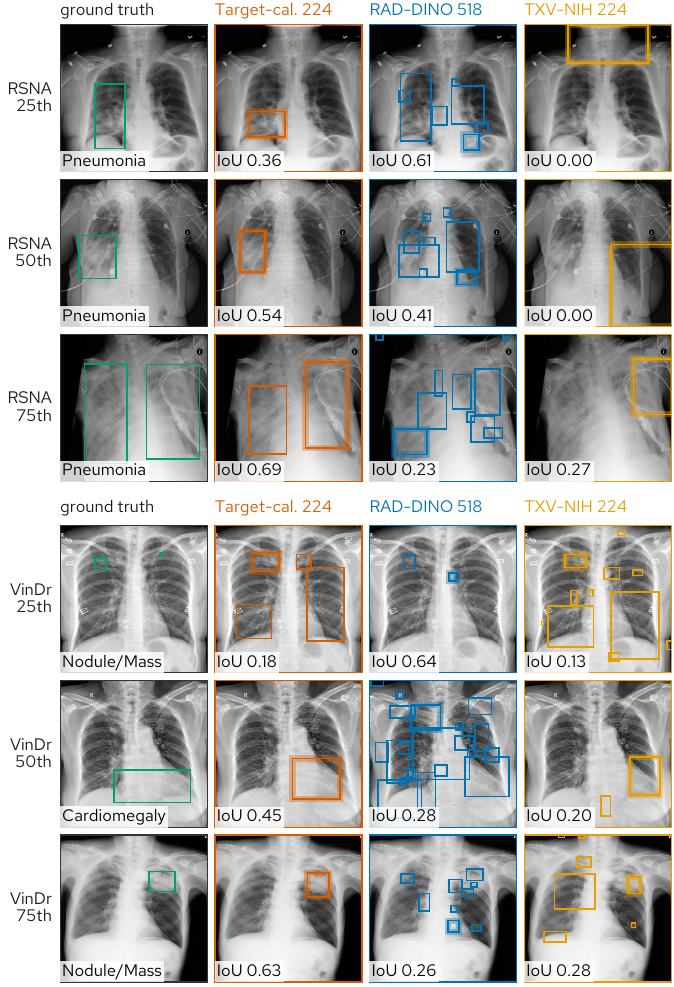}
\caption{Qualitative transfer examples at the 25th, 50th and 75th percentile of per-image CLARiTy target-calibrated IoU (224 px; no weight training on the target). RSNA is the NIH-test-derived split. Green: ground truth. CLARiTy Target-cal., 224 (vermilion), RAD-DINO 518 (blue) and TXV-NIH 224 (orange) show the boxes from their own OneBox extractors, with parameters fitted on the validation images. The box giving the printed IoU carries a light halo. The rows are selected by rank. For each corpus, the positive test images are ranked by their CLARiTy Target-cal., 224 per-image IoU (the mean over an image's findings of the best predicted-box IoU), with ties broken by a fixed seed. The image at each percentile is shown, taking the nearest-ranked image on which all three models store at least one box, so the rows span the range of CLARiTy target-calibrated outcomes rather than its best or worst cases.}
\label{fig:transfer-overlays}
\end{figure}

\Cref{tab:transfer_vindr_per_finding} gives VinDr-CXR image-level IoU accuracy for each finding and all six model settings at IoU thresholds $T = 0.1$ to $0.5$, \cref{tab:transfer_per_T_extra} the per-threshold accuracies for both corpora and all six settings, and \cref{tab:transfer_paired_extra} the paired differences for the refit and for RAD-DINO with their $p$-values. RSNA has a single finding, so its per-threshold values are that finding's. Per-finding values carry no intervals and rest on 18 to 309 positives, so the per-finding table is descriptive. At $T = 0.5$ the largest per-finding gap to RAD-DINO is on Cardiomegaly (309 positives), where the target-calibrated model at 512 px scores 0.926 against 0.019 for RAD-DINO 518, whereas on Pleural effusion RAD-DINO is ahead (0.072 against 0.054). Across findings the target-calibrated 512 model ranges from 0.926 on Cardiomegaly down to 0.054 on Pleural effusion and 0.056 on Pneumothorax, and the 224 model has the same highest- and lowest-scoring findings. The two calibrated settings do not order uniformly across findings: at $T = 0.5$ the 512 model is ahead of its 224 counterpart on four findings, behind it on Infiltration (0.241 against 0.293), and level on Pneumothorax. The macro row is therefore the appropriate summary of the comparison. \Cref{fig:transfer-overlays} shows three qualitative cases per corpus, selected by the rule stated in its caption.

\begin{table}[!t]
\centering
\caption{Per-threshold image-level IoU accuracy for both corpora and all six settings (point estimates; the 95\,\% bootstrap intervals are not reproduced). RSNA is the NIH-test-derived split and has one finding, so its row is that finding. The VinDr-CXR rows are the unweighted macro over six findings. The highest value for each corpus and threshold is shown in bold.}
\label{tab:transfer_per_T_extra}
\setlength{\tabcolsep}{0.8pt}
\footnotesize
\begin{tabular}{l l r r r r r}
\toprule
Corpus & Setting & $T{=}0.1$ & $T{=}0.2$ & $T{=}0.3$ & $T{=}0.4$ & $T{=}0.5$ \\
\midrule
\multirow{6}{*}{\shortstack[l]{RSNA\\(1 finding)}} & CLARiTy NIH-cal., 224 & \bfseries 0.9838 & 0.9073 & 0.7829 & 0.6423 & 0.5087 \\
 & CLARiTy NIH-cal., 512 & 0.9730 & 0.8733 & 0.7374 & 0.5925 & 0.4469 \\
 & CLARiTy Target-cal., 224 & 0.9784 & \bfseries 0.9289 & \bfseries 0.8277 & \bfseries 0.7053 & \bfseries 0.5670 \\
 & CLARiTy Target-cal., 512 & 0.9791 & 0.9054 & 0.7833 & 0.6454 & 0.4886 \\
 & RAD-DINO 518 & 0.9776 & 0.8853 & 0.6852 & 0.4396 & 0.2298 \\
 & TXV-NIH 224 & 0.1251 & 0.0398 & 0.0135 & 0.0027 & 0.0004 \\
\midrule
\multirow{6}{*}{\shortstack[l]{VinDr-CXR\\(6 findings)}} & CLARiTy NIH-cal., 224 & 0.6832 & 0.5138 & 0.4073 & 0.3226 & 0.1889 \\
 & CLARiTy NIH-cal., 512 & 0.7491 & 0.5671 & \bfseries 0.4511 & \bfseries 0.3576 & 0.2484 \\
 & CLARiTy Target-cal., 224 & 0.6840 & 0.5191 & 0.3992 & 0.3326 & 0.2514 \\
 & CLARiTy Target-cal., 512 & 0.7068 & 0.5454 & 0.4261 & 0.3495 & \bfseries 0.2851 \\
 & RAD-DINO 518 & \bfseries 0.7744 & \bfseries 0.6143 & 0.3556 & 0.1494 & 0.0734 \\
 & TXV-NIH 224 & 0.5368 & 0.3446 & 0.1956 & 0.0949 & 0.0306 \\
\bottomrule
\end{tabular}
\end{table}

\renewcommand{\theHtable}{\thesection.\arabic{table}}
\begin{table}[!t]
\centering
\caption{Paired per-threshold differences, comparison minus baseline ($\Delta$), for the RSNA (NIH-test-derived split) and VinDr-CXR test sets, with two-sided paired-bootstrap sign p-values from common resamples ($B\,=\,2{,}000$). The baseline is the CLARiTy NIH-calibrated setting for the CLARiTy target-calibrated rows and the CLARiTy target-calibrated setting for the RAD-DINO rows. A positive $\Delta$ means the comparison scores higher. $\Delta$ is the bootstrap-mean difference, not the point-estimate difference.}
\label{tab:transfer_paired_extra}
\setlength{\tabcolsep}{0.8pt}
\footnotesize
\begin{tabular}{l l r r r r r}
\toprule
Comparison & & $T{=}0.1$ & $T{=}0.2$ & $T{=}0.3$ & $T{=}0.4$ & $T{=}0.5$ \\
\midrule
\multicolumn{7}{l}{\textit{RSNA (NIH-test-derived split)}} \\
\multirow{2}{*}{\shortstack[l]{CLARiTy Target-cal., 224\\ $-$ CLARiTy NIH-cal., 224}} & $\Delta$ & $-$0.0053 & +0.0216 & +0.0448 & +0.0629 & +0.0583 \\
 & p & 0.0560 & $<$0.0005 & $<$0.0005 & $<$0.0005 & $<$0.0005 \\
\addlinespace
\multirow{2}{*}{\shortstack[l]{CLARiTy Target-cal., 512\\ $-$ CLARiTy NIH-cal., 512}} & $\Delta$ & +0.0061 & +0.0320 & +0.0463 & +0.0528 & +0.0416 \\
 & p & 0.0190 & $<$0.0005 & $<$0.0005 & $<$0.0005 & $<$0.0005 \\
\addlinespace
\multirow{2}{*}{\shortstack[l]{RAD-DINO 518\\ $-$ CLARiTy Target-cal., 224}} & $\Delta$ & $-$0.0008 & $-$0.0436 & $-$0.1424 & $-$0.2656 & $-$0.3377 \\
 & p & 0.8810 & $<$0.0005 & $<$0.0005 & $<$0.0005 & $<$0.0005 \\
\addlinespace
\multirow{2}{*}{\shortstack[l]{RAD-DINO 518\\ $-$ CLARiTy Target-cal., 512}} & $\Delta$ & $-$0.0016 & $-$0.0201 & $-$0.0981 & $-$0.2058 & $-$0.2595 \\
 & p & 0.7240 & 0.0050 & $<$0.0005 & $<$0.0005 & $<$0.0005 \\
\midrule
\multicolumn{7}{l}{\textit{VinDr-CXR}} \\
\multirow{2}{*}{\shortstack[l]{CLARiTy Target-cal., 224\\ $-$ CLARiTy NIH-cal., 224}} & $\Delta$ & +0.0008 & +0.0053 & $-$0.0080 & +0.0101 & +0.0629 \\
 & p & 0.8900 & 0.7230 & 0.2540 & 0.0990 & $<$0.0005 \\
\addlinespace
\multirow{2}{*}{\shortstack[l]{CLARiTy Target-cal., 512\\ $-$ CLARiTy NIH-cal., 512}} & $\Delta$ & $-$0.0419 & $-$0.0213 & $-$0.0255 & $-$0.0077 & +0.0367 \\
 & p & 0.0200 & 0.2740 & 0.1280 & 0.5350 & 0.0010 \\
\addlinespace
\multirow{2}{*}{\shortstack[l]{RAD-DINO 518\\ $-$ CLARiTy Target-cal., 224}} & $\Delta$ & +0.0903 & +0.0954 & $-$0.0437 & $-$0.1829 & $-$0.1778 \\
 & p & $<$0.0005 & 0.0010 & 0.1470 & $<$0.0005 & $<$0.0005 \\
\addlinespace
\multirow{2}{*}{\shortstack[l]{RAD-DINO 518\\ $-$ CLARiTy Target-cal., 512}} & $\Delta$ & +0.0670 & +0.0687 & $-$0.0703 & $-$0.1999 & $-$0.2111 \\
 & p & 0.0060 & 0.0150 & 0.0090 & $<$0.0005 & $<$0.0005 \\
\bottomrule
\end{tabular}
\end{table}

\renewcommand{\theHtable}{\thesection.\arabic{table}}
\FloatBarrier
\bibliographystyle{\bibstyleName}
\bibliography{db}

\end{document}